\documentclass[preprint,12pt]{elsarticle}

\usepackage{graphicx}
\usepackage{amssymb}

\usepackage{lineno}

\usepackage{graphicx}
\usepackage[utf8]{inputenc}
\usepackage{lmodern}
\usepackage[T1]{fontenc}
\usepackage{lscape}
\usepackage{amsmath}	
\usepackage{amssymb}
\usepackage{cases}
\usepackage{graphicx}
\usepackage{caption}
\usepackage{comment}

\usepackage{makecell}
\usepackage{multirow}
\usepackage{adjustbox}
\usepackage{amsfonts}
\usepackage{float} 
\usepackage{diagbox}
\usepackage{fullpage}
\usepackage{url}
\usepackage{url}
\usepackage{algorithm}
\usepackage{algpseudocode}
\usepackage{rotating}
\usepackage{eurosym}
\usepackage{wrapfig}
\usepackage[final]{pdfpages} 
\usepackage{epstopdf} 
\usepackage[a4paper]{geometry}
\usepackage[nottoc]{tocbibind}
\usepackage{placeins}
\usepackage{float}
\usepackage{listings}
\usepackage{tikz-cd}
\usepackage{color, colortbl}
\usepackage{hyperref}
\usepackage{xcolor}
\usepackage[printonlyused]{acronym}
\usepackage{graphicx, caption, subcaption}
\usepackage{booktabs}
\usepackage[normalem]{ulem}
\useunder{\uline}{\ul}{}
\usepackage{caption}  
\usepackage{sidecap}
\hypersetup{
colorlinks,
citecolor=black,
filecolor=black,
linkcolor=black,
urlcolor=black
} 
\begin{document}

\begin{frontmatter}


\title{Contrastive Learning Boosts Deterministic and Generative Models for Weather Data%
  \tnoteref{t1}}

\tnotetext[t1]{Adapted from an MSc thesis submitted by the same author to Imperial College London, 2025. 
The original thesis is available at \url{https://www.nathanbaileyw.com/msc-thesis}.}

\author{Nathan Bailey\corref{cor1}\fnref{fn1}}
\ead{nathanbaileyw@gmail.com}

\fntext[fn1]{Supervised by Dr Sibo Cheng, Faculty of Engineering, Imperial College London and Dr Cesar Quilodran-Casas, The Grantham Institute for Climate Change, Faculty of Natural Sciences, Imperial College London.}


\begin{abstract}

Weather data, comprising variables such as temperature, geopotential, and wind speed, poses significant challenges due to its high dimensionality and multimodal nature. Creating low-dimensional embeddings requires compressing this data into a compact, shared latent space. This compression is required to improve the efficiency and performance of downstream tasks, such as forecasting or extreme-weather detection. This has been highlighted more recently by the foundation model AlphaEarth \cite{alpha_earth}, which demonstrates the need for a pre-trained model capable of generating robust representations of high-dimensional weather and climate data. \\

Self-supervised learning, particularly contrastive learning, offers a way to generate low-dimensional, robust embeddings from unlabelled data, enabling downstream tasks when labelled data is scarce. Despite initial exploration of contrastive learning in weather data, particularly with the ERA5 dataset, the current literature does not extensively examine its benefits relative to alternative compression methods, notably autoencoders. Moreover, current work on contrastive learning does not investigate how these models can incorporate sparse data, which is more common in real-world data collection. It is critical to explore and understand how contrastive learning contributes to creating more robust embeddings for sparse weather data, thereby improving performance on downstream tasks. \\

Our work extensively explores contrastive learning on the ERA5 dataset, aligning sparse samples with complete ones via a contrastive loss term to create SPARse-data augmented conTRAstive spatiotemporal embeddings (SPARTA). We show that our method and experiments improve performance over an autoencoder across a range of downstream tasks. This advance is achieved by introducing a temporally aware batch sampling strategy and a cycle-consistency loss to improve the structure of the latent space. Furthermore, we propose a novel graph neural network fusion technique to inject domain-specific physical knowledge, thereby further enhancing our contrastive learning method. Ultimately, our results demonstrate that contrastive learning is a feasible and advantageous compression method for sparse geoscience data, thereby enhancing performance in downstream tasks. \\
\end{abstract}




\end{frontmatter}

\section*{Main Notations}
\begin{center}
\begin{tabular}{@{}ll@{}}
\toprule
\textbf{Notation} & \textbf{Description} \\
\midrule
$\mathcal{E}_x$ & Autoencoder Encoder \\
$\mathcal{D}_x$ & Autoencoder Decoder \\
$f(\cdot)$ & SimCLR Encoder \\
$g(\cdot)$ & SimCLR Projector \\
$\tau$ & SimCLR Loss Temperature Hyperparameter \\
$\mathcal{L}_{\text{r}}$ & Decoder Loss \\
$\ell_{\text{c}}$ & Intermediate Contrastive Loss \\
$\mathcal{L}_{\text{c}}$ & Total Contrastive Loss \\
$\mathcal{L}$ & Total Loss \\
$\ell_{\text{cycle}}$ & Cycle Loss \\
$\alpha$ & Joint Loss Function Term for Blending Training Stage \\
$k$ & Decay Hyperparameter for $\alpha$ \\
$\ell_{i,j}$ & Individual Loss for SimCLR pair $(i, j)$\\
$\mathbf{h}$ & Contrastive Embeddings (Latent Space) \\
$\mathbf{z}$ & Contrastive Projections \\
$\mathbf{x}$ & Ground Truth Data \\
$\mathbf{\hat{x}}$ & Reconstructed (Predicted) Data \\
\bottomrule
\end{tabular}
\end{center}

\section{Introduction}
\label{sec:intro}
In recent years, machine learning (ML) has been increasingly deployed for weather forecasting. Deep Learning models such as Pangu-Weather \cite{pangu} and GraphCast \cite{graph_cast} aim to replace or complement numerical weather prediction (NWP), a more traditional forecasting method. One of the main challenges with weather data is the sparsity of observations. This sparsity arises either from the high cost of collecting complete data or from sensor readings that provide only partial measurements \cite{sparse_2, sparse_5}. Sparsity poses significant challenges in machine learning applications, as samples are often unlikely to accurately reflect the underlying data distribution, leading to overfitting \cite{sparse_4}. \\ 

Weather data samples often reside in high dimensions. Due to the curse of dimensionality, these tasks become more challenging, as the sheer volume of data needed to prevent overfitting grows exponentially. Additionally, this increases the computational requirements for sample processing and model training. This compounds, as datasets often consist of multiple data variables (modes), substantially increasing the data's dimensionality, making downstream performance on raw data samples more difficult. Therefore, a unified model is needed that can reliably handle data sparsity and compress it into a shared low-dimensional latent space, enabling downstream tasks to run more efficiently and achieve higher performance. \\

One method to achieve this shared latent space is to compress the data using an autoencoder. However, an alternative method is contrastive learning, a type of self-supervised learning (SSL) that learns data representations by contrasting positive and negative samples. In this way, within the resulting reduced representation space, positive samples are pulled closer together, and negative samples are pushed apart, creating structure in the latent space \cite{intro_4}. Through this method, the model learns high-quality data embeddings that can be utilised in downstream tasks such as classification or segmentation. Contrastive learning performance can match or exceed that of supervised learning \cite{intro_4}, enabling the use of a range of datasets with sparse labels in tasks that would typically require extensive supervision. \\

Self-supervised learning and, by extension, contrastive learning therefore naturally lend themselves to weather analysis datasets, as these are often collected without labels. Therefore, contrastive learning provides a means to create significantly smaller, more structured representations of weather data, thereby reducing dimensionality and improving downstream performance. \\

Existing work has explored contrastive learning and general data compression for weather data; however, no extensive investigation of this topic has been conducted. \\

Wang et al \cite{con_era5_3} build a SimCLR model to classify weather systems in the latent space, demonstrating the strength of contrastive learning on weather datasets compared to traditional approaches. They also exhaustively evaluate various augmentation techniques using ERA5 data, proving that resizing and smoothing yield the best results. However, they only consider classifying weather systems using clustering or a linear classifier on the latent representations, with a subset of the data's latitude and longitude. Furthermore, an extensive evaluation using multiple downstream tasks and decoding back from the latent representation is not considered. \\

Similarly, Wang et al \cite{con_era5_1} consider contrastive learning on ERA5 data using a hybrid fusion method. However, similar to Wang et al, they optimise the model on a single downstream task: tropical cyclone prediction. \\

Han et al \cite{con_era5_4} create a VAE to compress the ERA5 dataset into a latent space, reducing its size and thereby widening access to it. They demonstrate that this method can achieve comparable forecasting performance on latent data to that of full dimensionality. Whilst this is an important first step, they do not consider the benefit that contrastive learning methods could bring. Furthermore, they do not decode the resulting latent predictions back to their original dimensionality, a step that is important for real-world applications. \\

The approaches highlighted consider only simple downstream tasks that do not require decoding the latent representations back to their original dimensionality. This limits the downstream tasks that can be utilised for evaluation. Tasks such as latent diffusion or forecasting require this decoding step to interpret the results. Moreover, they do not fully explore the available multimodal fusion methods. Contrastive learning approaches typically focus on a single data mode. However, weather data is inherently multimodal; therefore, to perform contrastive learning, it is necessary to fuse the data modalities. Evaluating available fusion methods is a useful and necessary step toward contrastive learning for the ERA5 dataset. \\

Finally, these methods treat the reanalysis dataset as-is and do not account for the sparsity or noise in weather data. Existing methods addressing sparsity in weather datasets \cite{sparse_1, sparse_2, sparse_3} focus on training masked autoencoders to reconstruct sparse data. While this is an important area of focus within weather datasets, none of these methods consider creating robust representations of sparse or noisy data within the latent space for downstream tasks. \\

Therefore, we can conclude that contrastive learning in geoscience applications, such as weather forecasting, has not been widely explored. We focus our work on this gap, exploring it in more detail to build an end-to-end model. Our model integrates multiple sparse modalities via fusion techniques to generate robust latent representations, termed SPARse-data augmented conTRAstive spatiotemporal embeddings (SPARTA). Furthermore, to demonstrate the benefits of contrastive learning, we benchmark all our experiments against a representative autoencoder. \\

Specifically, we make the following contributions:

\begin{enumerate}
    \item \textbf{We construct a end-to-end contrastive learning model}. Building on the literature, we integrate a decoder into the contrastive learning pipeline to create an end-to-end system that enables downstream predictions to be utilised in the input space. 
    \item \textbf{We highlight the benefits of contrastive learning to a multitude of downstream tasks}. To achieve our goal of creating a robust model and, consequently, a robust latent space, we evaluate our model on three downstream tasks: autoregressive forecasting, conditional latent diffusion, and latent classification. Through the use of a novel \textbf{hard negative sampling scheme and cycle consistency loss}, we demonstrate that contrastive learning yields a superior latent space for all tasks compared to a representative autoencoder.
    \item \textbf{We show that late-fusion offers substantial benefits over early-fusion}. We integrate late-fusion techniques into our model and demonstrate that our novel \textbf{graph neural network (GNN) approach} outperforms the state-of-the-art self-attention method.
\end{enumerate}

The remainder of this paper is as follows. Section \ref{sec:dataset} introduces the ERA5 dataset. The proposed machine learning methodology is introduced in section \ref{sec:method}. Finally, we perform a detailed evaluation in section \ref{sec:experiments}, thoroughly analysing the latent space and performing ablation tests on our subcomponents to highlight their performance gains. 

\section{Dataset} \label{sec:dataset}

Throughout our work, we utilise the ERA5 dataset. ERA5 is the fifth-generation European Centre for Medium-Range Weather Forecasting (ECMWF) reanalysis dataset, containing time-series data from 1959 onwards. A reanalysis dataset combines observations and model data to create a complete dataset \cite{era5}. It therefore provides consistent data for our experiments. Moreover, it has been used in early work on contrastive learning in geoscience \cite{con_era5_3, con_era5_1}. ERA5 consists of 62 data variables (modes) that span the complete longitude and latitude of the globe \cite{era5_2}. The ERA5 dataset can be used for a range of tasks, including weather forecasting, extreme event monitoring, and classification. \\

For our experiments, we use the following five variables: air temperature at 2 metres above the surface, geopotential, horizontal and vertical wind speeds at 100 metres above the Earth's surface, and specific humidity. All but temperature are associated with pressure levels, with the 850 hPa level used in accordance with \cite{con_era5_3}. We use the lowest-dimensional data, which corresponds to a resolution of 5.625° × 5.625° (64x32), which spans the full range of latitude and longitude, resulting in a dataset of size $X \in \mathbb{R}^{B \times 5 \times 64 \times 32}$, where $B$ is the batch size, and each variable is stacked as a channel. 

\section{Machine Learning Methodology} \label{sec:method}
\subsection{Model: Deterministic Autoencoder}
\vspace{3mm}

The fundamental goal of our work is to obtain a general embedding that is suitable for a variety of downstream tasks. One method to achieve this goal is to use an autoencoder. This is a type of neural network that aims to create a compressed representation by encoding and subsequently decoding data samples. The structure of an autoencoder consists of an encoder, denoted as \(\mathcal{E}_x\), that compresses the input vector ($\textbf{x}_{i}$) to a representation ($\textbf{h}_{i}$), often called the latent space, and a decoder, denoted as \(\mathcal{D}_x\), that maps this latent vector to a reconstruction of the input ($\hat{\textbf{x}}_{i}$). 

\begin{equation} \label{eq:simclr_arrow_1}
\begin{tikzcd}
\mathbf{x}_i \arrow[r, "\mathcal{E}_x"] & 
\mathbf{h}_i \arrow[r, "\mathcal{D}_x"] & 
\hat{\mathbf{x}}_i
\end{tikzcd}
\end{equation}

A deterministic autoencoder involves no stochasticity and learns to map the compressed representation back to the original data point. This type of autoencoder is optimised using a mean squared error (MSE) loss computed between the reconstructed and ground-truth data, as shown in equation \ref{eq:mse_loss} where $N$ is the size of the training dataset. Due to the model's deterministic nature, the same data point maps to the same compressed point in the latent space.

\begin{equation} \label{eq:mse_loss}
    \mathcal{L}_{\text{MSE}} = \frac{1}{N}\sum_{i=1}^{N}\left\| \hat{\textbf{x}}_{i} - \textbf{x}_{i}\right\|^2 = \frac{1}{N}\sum_{i=1}^{N}\left\| \mathcal{D}_x(\mathcal{E}_x(\mathbf{x_i})) - \textbf{x}_{i}\right\|^2 
\end{equation}

Neural networks with enhanced structures, such as convolutional neural networks (CNNs), have been extended for use in autoencoders. Convolutional autoencoders (CAE) extend upon feedforward autoencoders by providing the benefits of locality, invariance and a reduction in parameters that are inherent to CNNs. \\

In this study, we utilise a convolutional autoencoder as a benchmark for our more sophisticated contrastive learning method. Specifically, we utilise a single ResNet-18 encoder to jointly process all modes of a single data sample, providing a latent dimension of $1000$. Our decoder is constructed from five decoder blocks, each consisting of a transposed convolutional layer for upscaling, followed by a residual convolutional block with $3\times 3$ kernels and ReLU activation.  

\subsection{Model: SimCLR}
\vspace{3mm}

For our contrastive learning comparison, we utilised the popular SimCLR model \cite{simclr}. This methodology was initially used for computer vision (CV) applications and consists of an encoder network denoted as \(f(\cdot)\) used to produce embeddings $\mathbf{h}$ and a projection network denoted as \(g(\cdot)\) used to produce projections $\mathbf{z}$ (see equation \ref{eq:simclr_arrow}). 

\begin{equation} \label{eq:simclr_arrow}
\begin{tikzcd}
\mathbf{x}_i \arrow[r, "f"] & 
\mathbf{h}_i \arrow[r, "g"] & 
\mathbf{z}_i
\end{tikzcd}
\end{equation}

SimCLR takes a batch of data samples and, for each, creates a positive pair \((i, j)\) by augmenting the data sample twice. These are used to create intermediate embeddings (latent space) \(\mathbf{h}_{i}, \mathbf{h}_{j}\) and projections \(\mathbf{z}_{i}, \mathbf{z}_{j}\). Within the batch (of size $N$), SimCLR optimises the network using the NT-Xent loss, as shown in equation \ref{simclr}. This maximises the cosine similarity (denoted by $\text{sim}()$) between normalised positive projections and minimises the cosine similarity between normalised negative projections. This is computed across all positive pairs (both \((i, j)\) and \((j, i)\)) in a mini-batch and averaged across all \(2N\) pairs, as shown in equation \ref{simclr_loss_full}. SimCLR demonstrated that utilising embeddings derived from this contrastive learning approach in downstream tasks achieved performance comparable to that of a ResNet-50 approach on labelled data \cite{simclr}.

\begin{equation} \label{simclr}
\ell_{i,j} = -\log\frac{ \exp\left( \text{sim}(\mathbf{z}_i, \mathbf{z}_j)/\tau \right) }{ \sum_{k=1}^{2N} \mathbf{1}_{[k \neq i]} \exp\left( \text{sim}(\mathbf{z}_i, \mathbf{z}_k)/\tau \right)}
\end{equation}

\begin{equation} \label{simclr_loss_full}
\mathcal{L}_\text{simclr} = \frac{1}{2N} \sum_{\text{all positive pairs}} \ell_{i,j}
\end{equation}

SimCLR was chosen over other popular contrastive learning models for its simplicity and the explicit separation it induces in the latent space via positive and negative samples. For example, alternatives such as supervised contrastive learning \cite{supcon} introduce additional complexity because they require explicit labels. Barlow Twins \cite{barlow_twins} does not use positive and negative samples, meaning that SimCLR should lead to embeddings of different weather states that are better separated in the latent space, due to it explicitly pushing away distinct samples. \\

A key consideration for SimCLR is the temperature hyperparameter $\tau$ introduced in the loss function; a smaller value, below $1$, increases the similarity between negative samples. This encourages the model to push negative samples farther from positive samples, creating more diverse representations. Tuning the temperature hyperparameter is key, and Chen et al \cite{simclr} demonstrate that a tuned $\tau$ can help the model learn from hard negatives, that is, negatives that are closely similar to the data point. They find that a temperature parameter of $0.1$ yields the best performance. \\

To construct our SimCLR model, we utilise the same joint ResNet-18 encoder as in our deterministic autoencoder and build a small linear projection network comprising three linear layers, with batch normalisation using global statistics. This provides us with an embedding space of 1000 and a projection space of 128. As highlighted by Richemond et al. \cite{part_2_3}, batch normalisation harms the quality of the encoder embeddings in contrastive learning. Therefore, we replace all but the first layer with group normalisation layers to provide batch-size independent normalisation. We make the same change to our autoencoder to enable a fair comparison.

\begin{figure}[H]
    \centering
    \includegraphics[width=\textwidth]{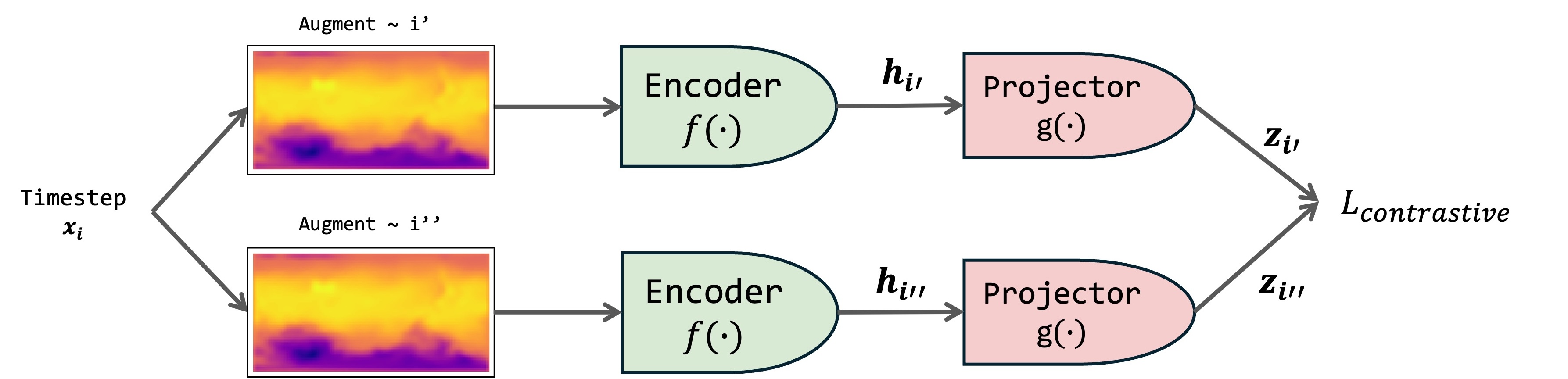}
    \caption{SimCLR Model Diagram}
    \label{fig:simclr}
\end{figure}

\subsection{Decoder Enhanced Contrastive Learning}
\vspace{3mm}
The utility of contrastive learning can be further expanded by jointly training a decoder on the contrastive embeddings. Typically, a decoder is not used with contrastive learning models, as they were initially developed for computer vision data, where the downstream tasks, such as classification, can be performed entirely in the latent space. However, in geoscience, downstream tasks such as forecasting require the latent-space results to be decoded for utilisation. In addition, it is essential to decode the embeddings to ensure that no information in the original space is lost in the latent space. Moreover, it has been shown that the representations and, therefore, performance can be enhanced by integrating a decoder into the model \cite{decoder_3, decoder_4}. As such, we train a decoder alongside the contrastive learning encoder. As contrastive learning approaches were not initially developed with an integrated decoder, this area of research is evolving. As such, several methods have been developed, which we summarise below.

\begin{itemize}
    \item The decoder can be trained alongside contrastive learning, blending the two losses in a form analogous to $L_{total} = \alpha L_{contrastive} + (1-\alpha) L_{decoder}$  \cite{decoder_3, decoder_4, decoder_5}. This can follow a pretraining approach, where the encoder is first trained solely on the contrastive objective. 
    \item The decoder can be trained alongside contrastive learning in two separate loops (inner and outer). In the outer loop, the contrastive objective is optimised, and in the inner loop, the autoencoder objective is optimised  \cite{decoder_1}. The outer loop uses a larger batch size, since contrastive learning methods often benefit from larger batch sizes. 
    \item Finally, the encoder can be split into a momentum and an online encoder. With the contrastive learning objective optimised across both encoders, the decoder reconstructs the online encoder's embeddings \cite{decoder_6}. 
\end{itemize}

There is also the decision of which data to reconstruct, since the encoder is trained on augmented data to optimise the contrastive objective. Kadeethum et al \cite{decoder_1} opt to reconstruct the non-augmented data, whereas Yao et al \cite{decoder_4} opt to reconstruct the augmented data. \\

We choose to adopt the first training approach, reasoning that the other two methods may be too destructive to the weights already learned by first pretraining the contrastive learning model. We opt to reconstruct masked non-augmented data, since this better aligns with the non-augmented data that the downstream tasks will encounter. 

\subsubsection{Sparse Observations}
\vspace{3mm}

As mentioned in section \ref{sec:intro}, the collected weather data is often sparse. During our work, we introduce sparse data into our training process as follows. \\

First, SimCLR requires the creation of positive pairs through data augmentation. We take advantage of this by forming a positive pair between a sparse and a complete sample. Specifically, in line with the work of \cite{con_era5_3}, we construct a positive pair by first resizing the sample to $160\times 80$, then randomly cropping it to $144\times 72$, and finally smoothing it with a $5\times5$ average filter. Then, for one sample in the pair, we randomly mask a percentage of its data points. We use a variable masking ratio in $[50\%, 90\%)$, sampled uniformly per sample. \\

As SimCLR was initially proposed for computer vision data, it takes a single data point and applies two augmentations to produce a positive pair. However, ERA5 is a time-series dataset, and as such, samples closer in time to a given point are inherently more similar (and thus more positive) than those farther away. Therefore, we follow the current literature \cite{time_2, time_3} and construct two positive pairs on either side of the timestep t, i.e. $[(t, t-\Delta t), (t, t+\Delta t)]$. We optimise the sum of the two contrastive losses for each pair, as shown in equations \ref{eq:p3_bach_eq_mod_2} and \ref{eq:p3_bach_eq_mod_3}, where $\mathcal{P}$ is the set of all $N$ positive pairs in the batch. $\Delta$ is selected in the range [1, 5] with a monotonically exponentially decreasing probability distribution, which is consistent with the initial contrastive work on the ERA5 data by Wang et al \cite{con_era5_3}. 

\begin{equation}
\label{eq:p3_bach_eq_mod_2}
\ell(t, \Delta t) = -\log \frac{
    \exp(\text{sim}(\mathbf{z}_{t}, \mathbf{z}_{t-\Delta t}) / \tau)
}{
    \sum_{k=1}^{2N} \mathbf{1}_{[k \ne t]} \exp(\text{sim}(\mathbf{z}_{t}, \mathbf{z}_{k}) / \tau)
} -\log \frac{
    \exp(\text{sim}(\mathbf{z}_{t}, \mathbf{z}_{t+\Delta t}) / \tau)
}{
    \sum_{k=1}^{2N} \mathbf{1}_{[k \ne t]} \exp(\text{sim}(\mathbf{z}_{t}, \mathbf{z}_{k}) / \tau)
    }
\end{equation}

\begin{equation} \label{eq:p3_bach_eq_mod_3}
\begin{aligned}
\ell_{\text{c}} = \frac{1}{2N} \sum_{(t, \Delta t) \in \mathcal{P}} \left( \ell(t, \Delta t) + \ell(\Delta t, t)  \right) 
\end{aligned}
\end{equation}

Secondly, when training the autoencoder and the decoder in the contrastive learning setting, we input a non-augmented sparse sample, using the same masking procedure as above, for reconstruction. This enables the decoder to learn to reconstruct complete data samples, similar to a masked autoencoder (MAE). 

\subsection{Contributions}
\vspace{3mm}

In addition to developing a more comprehensive contrastive learning model, we propose the following novel contributions to enhance our solution.

\subsubsection{Hard Negative Sampling Scheme}
\vspace{3mm}

Samples in a SimCLR batch are typically selected randomly. Whilst this would be adequate for image data, where all samples are equally negative with respect to one another, it is suboptimal for time-series data, where temporal proximity encodes meaningful structure. Therefore, we introduce explicit hard and soft negatives to the sampler during model training. Specifically, each sample (anchor) in the batch has a corresponding “hard” (within 30 timesteps of the anchor) and “soft” negative (outside 1000 timesteps of the anchor). \\

This approach mimics the effect of increasing the batch size in SimCLR, which increases the number of negative samples, making the separation task more challenging and producing higher-quality embeddings. We theorise that this approach similarly increases difficulty in the temporal domain, encouraging the model to learn more nuanced temporal distinctions. The model now leverages a similar set of temporally spaced samples in each batch, ensuring consistent tasks across batches and forcing the model to learn to separate hard and soft negatives in the latent space, resulting in higher-quality embeddings. In contrast, the previous sampling strategy left the range of negative samples to chance, potentially resulting in lower-quality embeddings.

\subsubsection{Cycle Consistency Loss}
\vspace{3mm}

Since we use two symmetric positive pairs around an anchor, this enables us to penalise the difference in distance between these points in the latent space. We formulate that better temporal smoothness in the latent space should boost downstream forecasting performance. Maintaining a constant distance between the embeddings of points that are the same timestep apart from an anchor should improve temporal smoothness. \\

Inspired by the work proposed by Xie et al \cite{smooth_1}, which proposes a jerk penalty using the third-order finite difference, we adapt this idea to our positive samples by proposing a penalty to penalise the second-order central finite difference between these positive points with the anchor. Mathematically, we suggest the following loss on our embeddings shown in equation \ref{eq:p3_cycle_loss}, denoting this as a cycle consistency loss. We add this to our contrastive loss, which is shown in equation \ref{p4:eq_c loss_1}.

\begin{equation}
\label{eq:p3_cycle_loss}
\ell_{cycle} = \frac{1}{N}\sum_{(t, \Delta t) \in \mathcal{P}}\frac{1}{d} \sum_{i=1}^{d} \left[ \mathbf{h}_{t+\Delta t}^i - 2\mathbf{h}_t^i + \mathbf{h}_{t-\Delta t}^i \right]^2
\end{equation}

\begin{equation} \label{p4:eq_c loss_1}
\mathcal{L}_{c} = \ell_{cycle} + \ell_{\text{c}}
\end{equation}

\subsection{Graph Neural Network Fusion Method} \label{sec:late_fusion}
\vspace{3mm}

As mentioned in section \ref{sec:intro}, current ERA5 methods that utilise contrastive learning do not extensively consider the various fusion methods that exist for multiple modalities. \\

In the simplest case, early-fusion fuses the modes of data by using a shared encoder for all modes \cite{tang}. Conversely, in late-fusion methods, an encoder is used for each mode, and the embeddings are combined before projection \cite{tang}, with the state of the art employing a self-attention fusion mechanism. Finally, a hybrid approach combines the two fusion mechanisms, which can be more flexible but adds additional complexity to model training \cite{tang}.  \\

In current literature, \cite{con_era5_3} only considers an early-fusion method, whereas \cite{con_era5_1} considers a hybrid fusion method. In the methodology detailed above, we utilised only an early-fusion method; we now also include late-fusion for ERA5. \\

\begin{figure}[H]
    \centering
    \includegraphics[width=\textwidth]{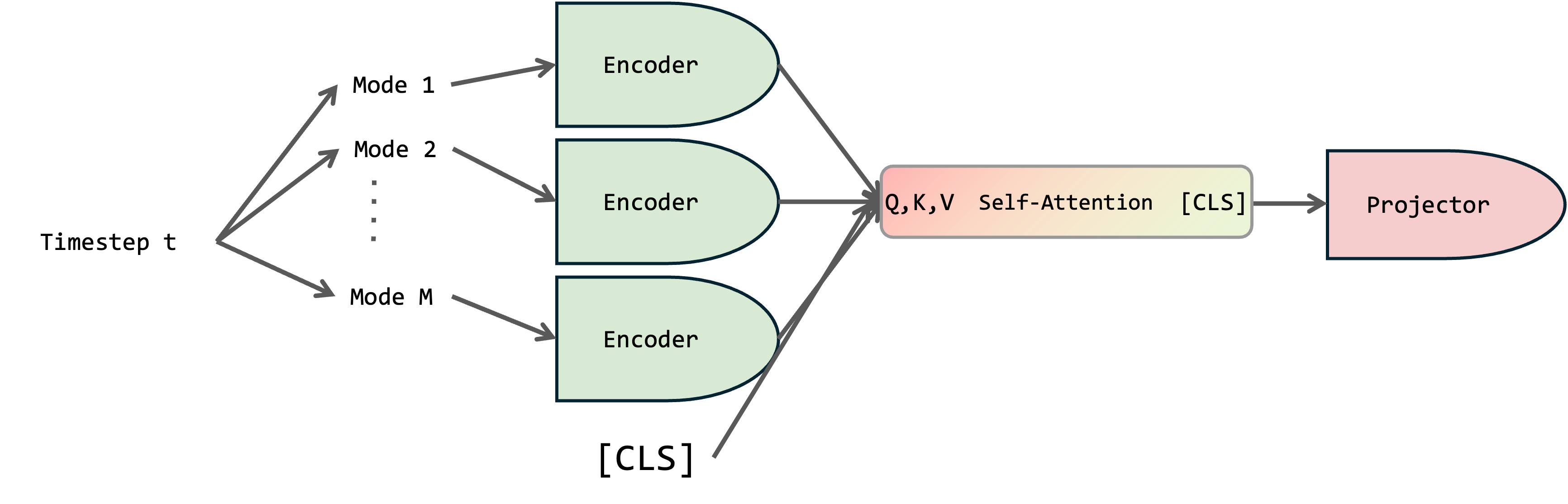}
    \caption{Self Attention Fusion Method}
    \label{fig:attn_fusion}
\end{figure}

In the self-attention late-fusion method, which is shown in figure \ref{fig:attn_fusion}, each embedding for each mode is input into a self-attention block as a token, enabling each modal embedding to interact with the others. Typically, a `[CLS]' token is included and used as the fused embedding, which serves as an average of all tokens in the block. Through the attention mechanism, the model can dynamically focus on specific modes, enabling better handling of noise and uncertainty in multimodal data \cite{fusion_2}. \\

\begin{figure}[H]
    \centering
    \includegraphics[width=\textwidth]{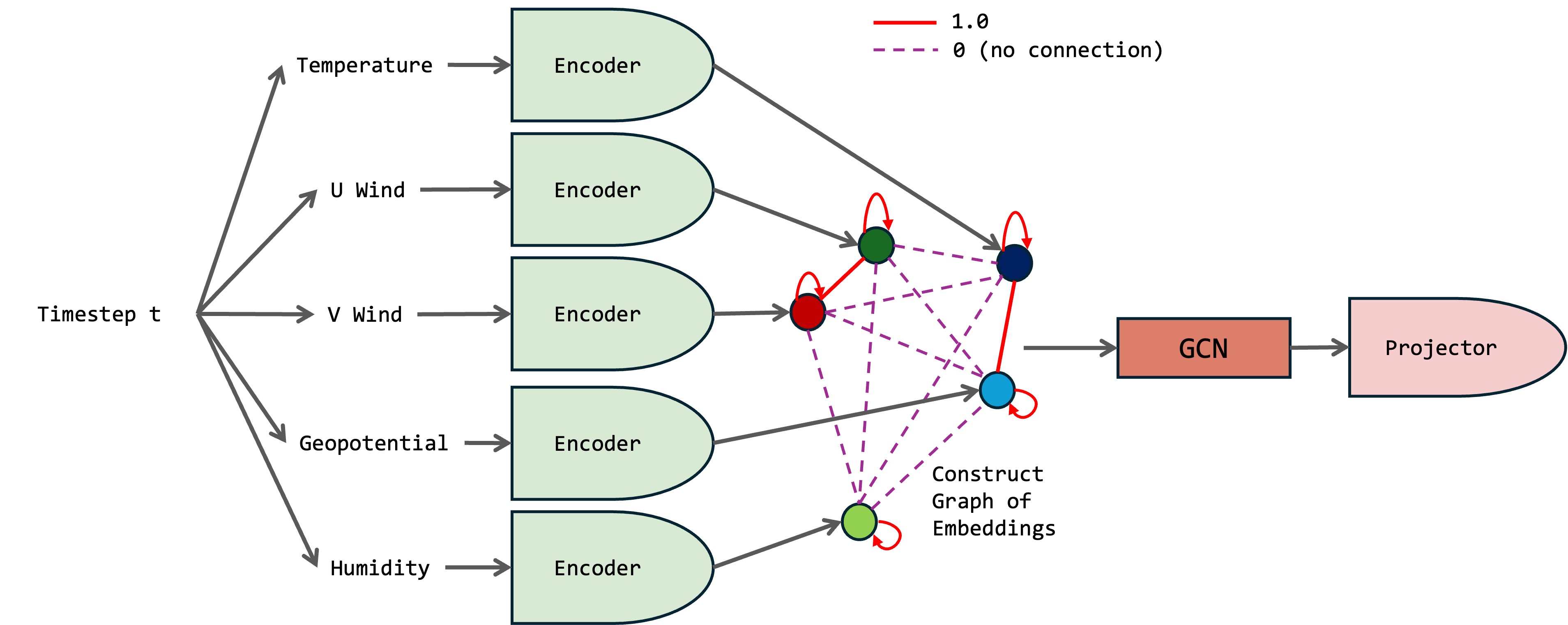}
    \caption{GNN Fusion Method}
    \label{fig:gnn}
\end{figure}

The self-attention method enables the creation of an embedding for each mode, with dynamic weighting across all modes. However, this becomes increasingly complex, as the weights are dynamically learned per sample, shifting the onus on the model to determine how best the modal embeddings should interact. To reverse this interaction decision, we introduce a novel graph neural network (GNN) methodology to fuse the embeddings, which we show in figure \ref{fig:gnn}. Graph neural networks, specifically graph convolutional networks (GCNs), were first introduced by Kipf et al. \cite{gnn_1} and enable data samples to interact with one another through a defined graph structure, represented by an adjacency matrix and through repeated neural network layers. We extend the application of GCNs to build a sparse graph of embeddings, noting that both wind modes will be highly correlated, as will the temperature and geopotential modes. We construct the adjacency matrix so that each mode has a self-connection, and only the chosen modes can interact. We then pass this graph through a GCN with graph-attention layers, allowing the weights between interacting modes to be learned dynamically for each sample. Since we are working with a graph with at most one hop, only one layer is required for a node to reach its neighbours. Employing multiple layers introduces nonlinearity into the network. However, it would mean the node representations would suffer from oversmoothing, a phenomenon that occurs when too many layers are present \cite{oversmoothing}. The final GCN embeddings are averaged to produce the final latent space embedding, which is then fed into the projection network. \\

We contrast our novel GNN approach with a self-attention model with 8 heads, in which each modal embedding is treated as a token, and a `[CLS]' token is used to obtain a fused latent space representation for the projection network. To distinguish between the modes, we additionally utilise learned modal embeddings, which are added to each token. 

\section{Downstream Tasks}
\vspace{3mm}

Once trained, the resulting latent space of both our autoencoder and contrastive learning models can be used for downstream tasks. Performing downstream tasks in the latent space rather than the original high-dimensional space offers a significant computational efficiency advantage. We evaluate our models using the following tasks, detailed below. For all, we first freeze both the encoder and decoder of our model, so that during training, no weight updates are applied, enabling us to compare and contrast approaches accurately.

\subsection{Autoregressive Forecasting}
\vspace{3mm}

\begin{figure}[H]
    \centering
    \includegraphics[width=\textwidth]{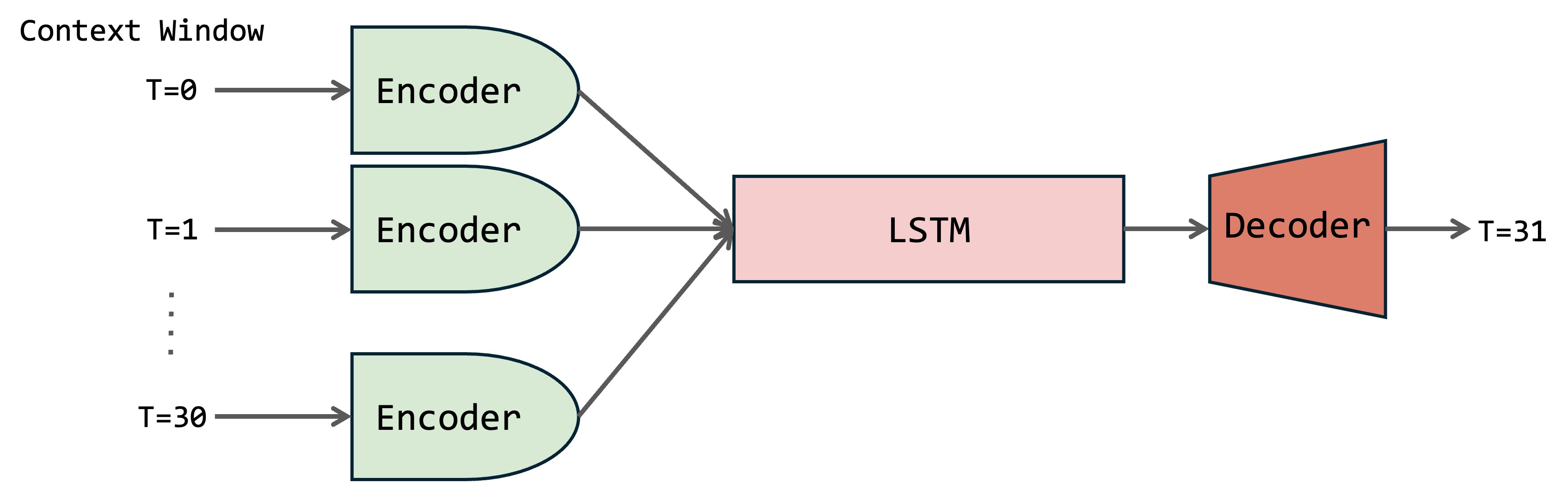}
    \caption{Forecasting Task}
    \label{fig:forecasting}
\end{figure}

For forecasting, we construct a simple sequence-to-sequence (seq2seq) encoder-decoder LSTM model in the latent space, as shown in figure \ref{fig:forecasting}. To perform the forecasting, a set of timesteps, labelled the look-back period (LB), masked with a 70\% ratio, are embedded ($\textbf{h}_{\text{cw}})$ and used as the input to the seq2seq encoder ($\textbf{e} = 
\mathcal{E}_\text{LSTM}(\textbf{h}_{\text{cw}})$). The decoder of the seq2seq model predicts the latent embeddings of the future timesteps $(\hat{\textbf{h}}^t = \mathcal{D}_\text{LSTM}(\hat{\textbf{h}}^{t-1}, c_{t-1})$), where $c_{t-1}$ is the hidden states of the LSTM for the previous timestep. These are then decoded to produce the forecast in the original dimensionality $(\hat{\textbf{x}}^t = \mathcal{D}(\hat{\textbf{h}}^t))$. During training, we predict only the embedding of the next timestep from the look-back window and minimise the MSE loss between the ground truth and the prediction, as shown in equation \ref{eq:forecast}. During evaluation, we generate the next 100 timesteps autoregressively, decode the results, and compare them against the ground truth. 

\begin{equation}
\label{eq:forecast}
L = \frac{1}{N}\sum_{i=1}^{N}\left\| \hat{\textbf{h}}_{i}^{cw+1} - \textbf{h}_{i}^{cw+1}\right\|^2
\end{equation}

\subsection{Conditional Latent Diffusion}
\vspace{3mm}

Diffusion models were first introduced by Ho et al \cite{diffusion_1}. A class of generative models, they produce new images by denoising a signal over a series of time steps \cite{generative_deep}[p208]. \\

To achieve this, data is noised by adding a small amount of Gaussian noise over a fixed number of timesteps $t$. Specifically, given an image $\mathbf{x}_{0}$, the noised image at time $t$ is given by $\sqrt{\bar{\alpha}_t} \, \mathbf{x}_0 + \sqrt{1 - \bar{\alpha}_t} \, \boldsymbol{\epsilon}$. Where a diffusion schedule gives $\bar{\alpha}_t$.  A neural network, typically a U-Net, is then used to predict the added noise in the image, using an MSE loss. Given the trained network, a noisy signal is transformed into an image over a fixed number of timesteps by undoing the noise the model predicts. \\

Rombach et al \cite{diffusion_2} extend upon this original work by performing diffusion in the latent space. During sampling, a latent code is generated and then decoded back to the full space to produce the sample. They further extend their work by introducing conditional diffusion models. These aim to condition the generation on a given input, such as text or representations. The encoded condition is used in cross-attention modules as keys and values, with queries derived from the hidden states that are inputs to the U-Net blocks. \\

\begin{figure}[H]
    \centering
    \includegraphics[width=\textwidth]{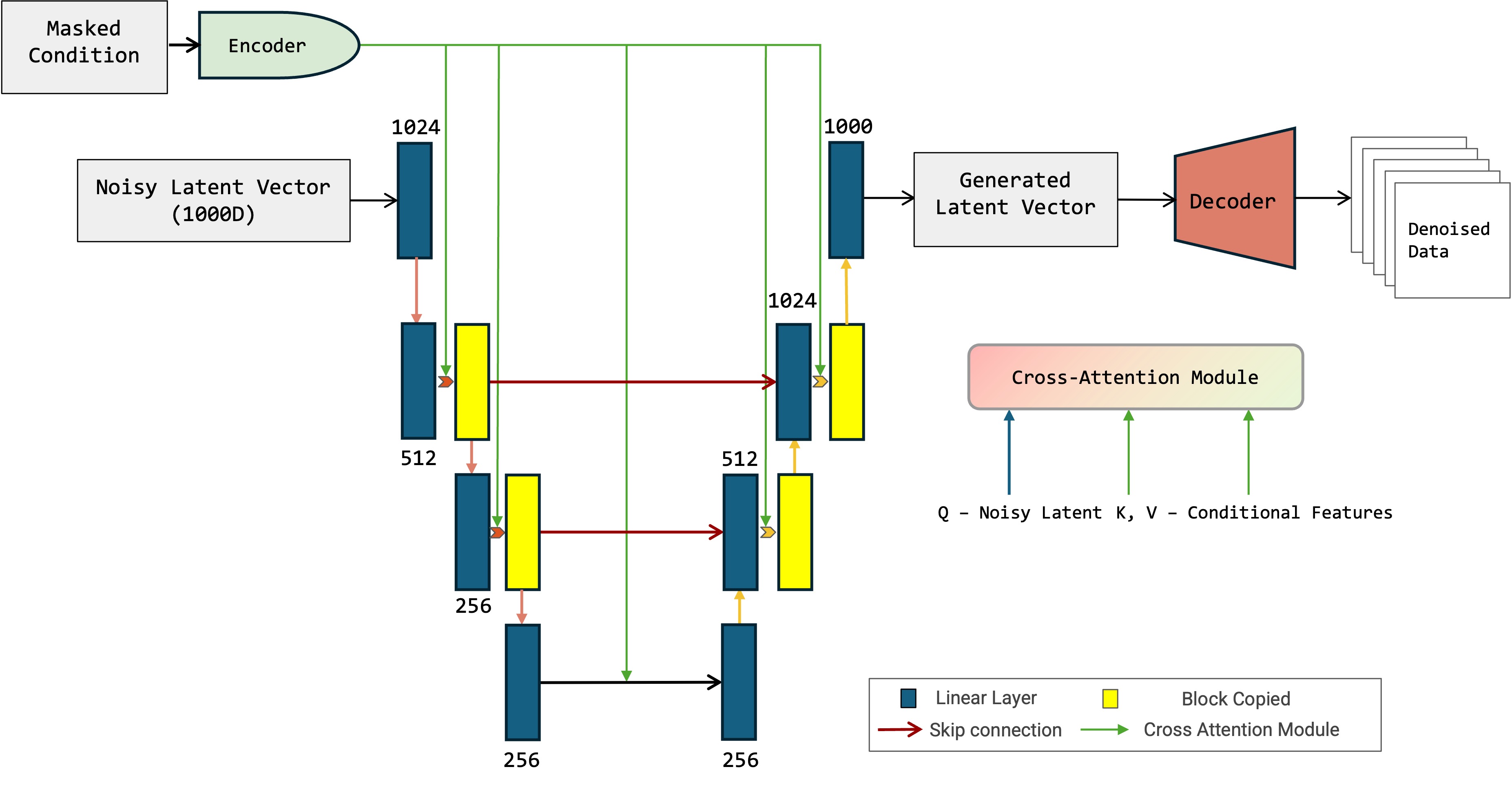}
    \caption{Conditional Latent Diffusion U-Net Architecture}
    \label{fig:latent_diff}
\end{figure}

We construct a conditional latent diffusion model ($\epsilon_\theta$) for evaluation, as shown in figure \ref{fig:latent_diff}. Implementing latent diffusion is particularly important for ERA5 data, as the data's high dimensionality makes training a full diffusion model substantially more complex. \\

For our diffusion model, we implement a linear U-Net in the latent space and condition each block on a 70\% masked data point ($\textbf{h}_\text{mask}$) via cross-attention. Conditioning on a sparse data sample should yield a reconstruction of the full sample, thereby providing an alternative way to recover it. We highlight our inference process in algorithm \ref{algo:diff}.

\begin{algorithm}[H]
\caption{Conditional Latent Diffusion Inference}
\label{algo:diff}
\begin{algorithmic}[1]
\Require Trained model $\epsilon_\theta$, conditioning $\textbf{h}_\text{mask}$, noise schedule $\{\alpha_t\}_{t=1}^T$
\State Sample initial latent: $\textbf{h}_T \sim \mathcal{N}(0, I)$
\For{$t = T$ \text{...} $1$}
    \State Predict noise: $\hat{\epsilon}_t \gets \epsilon_\theta(\textbf{h}_t, t, \textbf{h}_\text{mask})$
    \State Denoise latent: 
    \[
    \textbf{h}_{t-1} \gets \frac{1}{\sqrt{\alpha_t}} \Big(\textbf{h}_t - \frac{1-\alpha_t}{\sqrt{1-\bar{\alpha}_t}} \hat{\epsilon}_t \Big) + \sigma_t \, \eta_t, \quad \eta_t \sim \mathcal{N}(0, I)
    \]
\EndFor
\State Decode latent to data: $\textbf{x}_\text{gen} \gets D(\textbf{h}_0)$
\end{algorithmic}
\end{algorithm}

\subsection{Latent Classification}
\vspace{3mm}

Finally, we implement a classification model in the latent space to perform the inverse problem of season identification. In this task, given a masked data sample, the model will classify it into a season: `winter (0)', `spring (1)', `summer(2)', `autumn (3)' (see equation \ref{eq:classification_1}). To achieve this, we implement a small linear neural network ($\text{NN}_{\phi}$) with a classification head in the latent space and input test data, masked at variable ratios of [20\%, 90\%). As per classification tasks, we optimise the cross-entropy loss as shown in equation \ref{eq:classification_2}.

\begin{equation}
\label{eq:classification_1}
\textbf{s}_{i} = \text{NN}_{\phi}(\textbf{h}_{i})
\end{equation}

\begin{equation}
\label{eq:classification_2}
L_{\text{CE}} = - \frac{1}{N} \sum_{n=1}^{N} \sum_{i=1}^{K} y_{i}^n 
\log \frac{\exp(s_{i}^n)}{\sum_{j=1}^{K} \exp(s_{j}^n)}
\end{equation}

\section{Training Pipeline} \label{sec:train}
\vspace{3mm}

\begin{figure}[H]
    \centering
    \includegraphics[width=\textwidth]{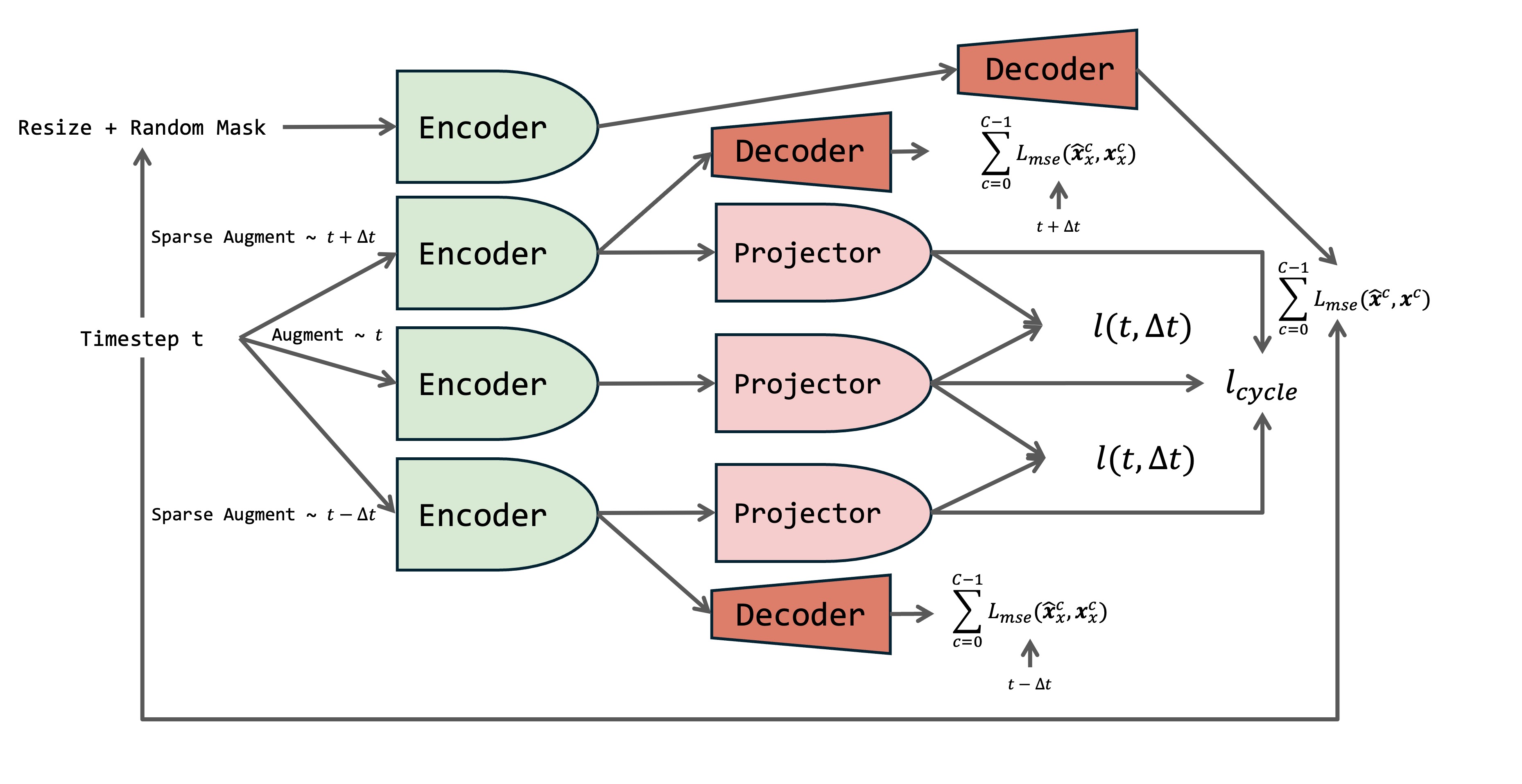}
    \caption{Early-Fusion SPARTA Model Diagram}
    \label{fig:diagram}
\end{figure}

We summarise our final training pipeline for all SPARTA methods below, comprising the following stages. Additionally, we show the early-fusion final model diagram in figure \ref{fig:diagram}. The late-fusion methods follow the same diagram, utilising the output of the GCN or self-attention block in place of the single encoder. 

\begin{enumerate}
    \item Pretraining the encoder on the contrastive objective ($\ell_{\text{c}}$) for 100 Epochs, where $\mathcal{P}$ is the set of all $N$ positive pairs in the batch.
    \item Blending in the decoder and the autoencoder objective by decaying alpha in our loss term. We start ($\alpha_{\text{start}}$) at $1.0$ and use an exponential decay to decrease the value: $\alpha(\text{epoch}) = \alpha_{\text{start}} \cdot e^{-k \cdot \text{epoch}}$, setting $k$ to $0.01$. We train this model for 250 epochs with the following loss: $\mathcal{L} = \alpha\mathcal{L}_{\text{c}} + (1-\alpha)\mathcal{L}_{\text{r}}$. For the reconstruction objective $\mathcal{L}_{\text{r}}$, we mask the non-augmented data sample, enabling the decoder to reconstruct the full data. Additionally, we train the decoder to reconstruct the augmented masked points at $t-\Delta t$ and $t+\Delta t$, as shown in equation \ref{p4:eq_c loss}, to enable more detailed embeddings when optimising the contrastive objective. 
    \item Freezing the encoder and training the decoder on the frozen embeddings for 200 epochs to reconstruct masked data. We optimise using only the reconstruction loss: $\mathcal{L}_{\text{r}}$.
    \item Training each downstream model. 
    \begin{itemize}
        \item For the forecasting model, we optimise the MSE loss between the latent forecast and the latent ground truth and train for 100 epochs.
        \item For the conditional latent diffusion model, we optimise the MSE loss between predicted and ground truth added noise, training for 300 epochs.
        \item For the latent classification model, we optimise the cross-entropy loss between the outputs and ground truth labels over 100 epochs.
    \end{itemize}
\end{enumerate}

\begin{equation}
\label{eq:p3_bach_eq_mod}
\ell(t, \Delta t) = -\log \frac{
    \exp(\text{sim}(\mathbf{z}_{t}, \mathbf{z}_{t-\Delta t}) / \tau)
}{
    \sum_{k=1}^{2N} \mathbf{1}_{[k \ne t]} \exp(\text{sim}(\mathbf{z}_{t}, \mathbf{z}_{k}) / \tau)
} -\log \frac{
    \exp(\text{sim}(\mathbf{z}_{t}, \mathbf{z}_{t+\Delta t}) / \tau)
}{
    \sum_{k=1}^{2N} \mathbf{1}_{[k \ne t]} \exp(\text{sim}(\mathbf{z}_{t}, \mathbf{z}_{k}) / \tau)
    }
\end{equation}

\begin{equation}
\label{eq:p4_cycle_loss}
\ell_{cycle} = \frac{1}{N}\sum_{(t, \Delta t) \in \mathcal{P}}\frac{1}{d} \sum_{i=1}^{d} \left[ \mathbf{h}_{t+\Delta t}^i - 2\mathbf{h}_t^i + \mathbf{h}_{t-\Delta t}^i \right]^2
\end{equation}

\begin{equation} \label{eq:simclr_time}
\begin{aligned}
\ell_{\text{c}} = \frac{1}{2N} \sum_{(t, \Delta t) \in \mathcal{P}} \left( \ell(t, \Delta t) + \ell(\Delta t, t)  \right) + \ell_{\text{cycle}}
\end{aligned}
\end{equation}

\begin{equation} \label{p4:eq_c loss}
\mathcal{L}_{c} = \ell_{\text{c}} + \sum_{x \in \{t-\Delta t, t+\Delta t\}}\sum_{c=0}^{C-1} \mathcal{L}_{\text{mse}}\left( \mathbf{\hat{x}}_{x}^c, \mathbf{x}^c_{x} \right)
\end{equation}

\begin{equation} \label{p4:eq_mse loss}
\mathcal{L}_{\text{r}} = \sum_{c=0}^{C-1} \mathcal{L}_{\text{mse}}\left( \mathbf{\hat{x}}^c, \mathbf{x}^c \right)
\end{equation}

This methodology is a significant advancement over current work. As mentioned in chapter \ref{sec:intro}, current contrastive learning approaches for downstream tasks on ERA5 data only train an encoder and evaluate solely in the latent space \cite{con_era5_3, con_era5_1}. Our method constructs a complete encoder-decoder framework, enabling downstream tasks such as forecasting that require decoding predictions. We also introduce sparsity, which none of the previous ERA5 contrastive learning methods considered. This significantly increases the utility of our model, as it better reflects how observations would be collected. Our method now enables sparse observations to be utilised directly in the latent space for downstream tasks, thereby further enhancing our framework's utility.

\section{Experimentation} \label{sec:experiments}
\subsection{Experimental Setup}
\vspace{3mm}

\subsubsection{Dataset Setup}
\vspace{3mm}
Through all our experiments, we utilise the ERA5 dataset as outlined in section \ref{sec:dataset}. We utilise all available samples from 1959 to 2023, allocating 60\% to training, 20\% to validation, and the remaining 20\% to testing. Data is split chronologically, using approximately 1959-1997 for training, 1998-2019 for validation, and 2010-2023 for testing. The testing data is used to train the downstream task, and we split it into training, validation, and test datasets in the same manner.

\subsubsection{Model Training}
\vspace{3mm}
To ensure a fair comparison, we train both our benchmark autoencoder and our SimCLR models with a batch size of 128. Additionally, we utilise the same hard negative sampling for both the autonecoder and the SimCLR models. Given that the cycle consistency loss was proposed as a direct result of the pairs generated for the SimCLR loss, its application to the autoencoder loss is unwarranted. \\

We found that the autoencoder performed best with a starting learning rate of $10^{-3}$ and trained the model for 180 epochs, optimising the MSE loss. Our contrastive learning models performed best at a learning rate of $10^{-4}$. In all models, we utilised a learning rate scheduler that reduced the learning rate by a factor of 10 when the validation loss plateaued for 10 epochs. Finally, for the contrastive learning models, we set the temperature hyperparameter to $0.3$. 

\subsubsection{Model Evaluation} \label{sec:model_eval}
\vspace{3mm}

Our downstream models are all trained with a batch size of 64 and a learning rate of $1e^{-2}$. \\

Specifically, for the forecasting task, we train three different models with different sampling intervals (S) and look-back windows (LB): LB 30 + S 1, LB 5 + S 5, LB 5 + S 10. We evaluate all models on the test set in an autoregressive manner, predicting the next 100 timesteps, using three metrics: RRMSE, SSIM and PSNR for all modes jointly. For models with a sampling interval greater than 1, we note that this significantly reduces the amount of training data. To negate this issue, we train these models using three different seeds and average the results. \\

For the diffusion task, we evaluate our model on the test data by measuring the RRMSE of the decoded data. To do this, we generate 20 batches and take the mean of the samples for decoding. This increases the vigour of our evaluation and allows us to compute the standard deviation of the generated latent data points, thereby enabling us to measure diffusion robustness. In addition, we also measure the realism and latent density scores in the embedding space. Latent density, proposed by Xu et al \cite{latent_gen_1}, is shown in equation \ref{eq:latent_gen_1}, where $\mathbf{h}_g$ is the generated latent point and $\mathcal{H}$ is the set of ground truth latent points. This score measures the average Gaussian-kernelised Euclidean distance between the generated sample and each training sample, with higher scores indicating higher-quality generated samples. For our experiments, as per the original work, we set $\sigma$ to 20. The realism score, shown in equation \ref{eq:latent_gen_2}, where $\mathrm{NN}_k$ is the kth nearest neighbour of $\mathbf{h}_i$ in $\mathcal{H}$, was initially proposed by Kynkäänniemi et al \cite{latent_gen_2}. This measures how closely the generated sample aligns with the image manifold. The higher the score, the closer the generated image is to the image manifold. Kynkäänniemi et al. utilised feature vectors from the generated images using a pre-trained VGG network. Since we do not have access to a pre-trained network, we use the latent vectors as is and set $k$ to 5. Therefore, our adaptation of the realism score quantifies the alignment between the generated latent codes and our learned manifold. 

\begin{equation} \label{eq:latent_gen_1}
D(\mathbf{h}_g, \mathcal{H}) = \frac{1}{|\mathcal{H}|} \sum_{\mathbf{h}_i \in \mathcal{H}} \exp\left(-\frac{\|\mathbf{h}_g - \mathbf{h}_i\|^2}{2\sigma^2}\right),
\end{equation}

\begin{equation} \label{eq:latent_gen_2}
R(\mathbf{h}_g, \mathcal{H}) = \max_{\mathbf{h}_i} \left\{ 
\frac{\|\mathbf{h}_i - \mathrm{NN}_k(\mathbf{h}_i, \mathcal{H})\|_2}
     {\|\mathbf{h}_g - \mathbf{h}_i\|_2}
\right\}.
\end{equation}

Finally, for the latent classification task, we evaluate our model on the test data and compute the cross-entropy metric. 

\section{Results}
\subsection{Early-Fusion} \label{sec:res}

\subsubsection{Forecasting}

\begin{table}[H]
\caption{Autoregressive Results - 100 Steps LB 30, S 1}
\label{table:res_1}
\centering
\resizebox{\textwidth}{!}{%
\begin{tabular}{|l|cccc|}
\hline
                                              & \multicolumn{4}{c|}{{\ul \textbf{Look-Back Window 30, Sampling Interval 1}}}                                                                                                                                \\ \hline
\multicolumn{1}{|c|}{\textbf{}}               & \multicolumn{1}{c|}{\textbf{RRMSE (\%) T=25}} & \multicolumn{1}{l|}{\textbf{RRMSE (\%) T=50}} & \multicolumn{1}{l|}{\textbf{RRMSE (\%) T=75}} & \multicolumn{1}{l|}{\textbf{RRMSE (\%) T=100}} \\ \hline
\textbf{SPARTA - Early-Fusion HN + Cycle}     & \multicolumn{1}{c|}{\textbf{15.71}}           & \multicolumn{1}{c|}{\textbf{16.45}}           & \multicolumn{1}{c|}{\textbf{16.84}}           & \textbf{17.04}                                 \\ \hline
\textbf{Autoencoder}                          & \multicolumn{1}{c|}{19.73}                    & \multicolumn{1}{c|}{20.13}                    & \multicolumn{1}{c|}{20.09}                    & 20.15                                          \\ \hline
\end{tabular}
}
\end{table}

\begin{table}[H]
\caption{Autoregressive Results - 100 Steps LB 5, S 5}
\label{table:res_2}
\centering
\resizebox{\textwidth}{!}{%
\begin{tabular}{|l|cccc|}
\hline
                                              & \multicolumn{4}{c|}{{\ul \textbf{Look-Back Window 5, Sampling Interval 5}}}                                                                                                                                 \\ \hline
\multicolumn{1}{|c|}{\textbf{}}               & \multicolumn{1}{c|}{\textbf{RRMSE (\%) T=25}} & \multicolumn{1}{l|}{\textbf{RRMSE (\%) T=50}} & \multicolumn{1}{l|}{\textbf{RRMSE (\%) T=75}} & \multicolumn{1}{l|}{\textbf{RRMSE (\%) T=100}} \\ \hline
\textbf{SPARTA - Early-Fusion HN + Cycle}     & \multicolumn{1}{c|}{14.89}                    & \multicolumn{1}{c|}{17.30}                    & \multicolumn{1}{c|}{17.74}                    & 17.71                                          \\ \hline
\textbf{Autoencoder}                          & \multicolumn{1}{c|}{16.51}                    & \multicolumn{1}{c|}{17.02}                    & \multicolumn{1}{c|}{17.13}                    & 17.30                                          \\ \hline
\end{tabular}
}
\end{table}

\begin{table}[H]
\caption{Autoregressive Results - 100 Steps LB 5, S 10}
\label{table:res_3}
\centering
\resizebox{\textwidth}{!}{%
\begin{tabular}{|l|cccc|}
\hline
                                              & \multicolumn{4}{c|}{{\ul \textbf{Look-Back Window 5, Sampling Interval 10}}}                                                                                                                                \\ \hline
\multicolumn{1}{|c|}{\textbf{}}               & \multicolumn{1}{c|}{\textbf{RRMSE (\%) T=25}} & \multicolumn{1}{l|}{\textbf{RRMSE (\%) T=50}} & \multicolumn{1}{l|}{\textbf{RRMSE (\%) T=75}} & \multicolumn{1}{l|}{\textbf{RRMSE (\%) T=100}} \\ \hline
\textbf{SPARTA - Early-Fusion HN + Cycle}     & \multicolumn{1}{c|}{13.62}                    & \multicolumn{1}{c|}{14.11}                    & \multicolumn{1}{c|}{14.75}                    & 15.03                                          \\ \hline
\textbf{Autoencoder}                          & \multicolumn{1}{c|}{18.83}                    & \multicolumn{1}{c|}{19.28}                    & \multicolumn{1}{c|}{19.05}                    & 19.05                                          \\ \hline
\end{tabular}
}
\end{table}

From tables \ref{table:res_1}, \ref{table:res_2}, and \ref{table:res_3}, we see that our early-fusion SPARTA model outperforms the baseline autoencoder on the latent forecasting task. To understand why, we compute the following metrics to measure the smoothness of the latent space.

\begin{enumerate}
    \item Based on work by \cite{smooth_3}, we determine the temporal smoothness of our latent space by computing the squared Euclidean distance between adjacent points in the latent space: $\Delta \mathbf{h}_t = \|\mathbf{h}_t - \mathbf{h}_{t-1}\|_2^2$. This provides a first-order smoothness measure, analogous to velocity. 
    \item Based on work by \cite{smooth_5}, we evaluate the cycle distance of points. This is analogous to acceleration and is achieved by computing the second-order central finite difference for embeddings: $\|\mathbf{h}_{t-1} - 2\mathbf{h}_t + \mathbf{h}_{t+1}\|_2^2$.
\end{enumerate}

For all metrics, we evaluate our model by computing them on a batch of test embeddings with a batch size of 256. This enables us to plot the metrics and determine the overall trend of the data, as well as the mean and standard deviation. \\

\begin{figure}[H]
    \centering
    \begin{subfigure}[b]{0.48\textwidth}
        \includegraphics[width=\textwidth]{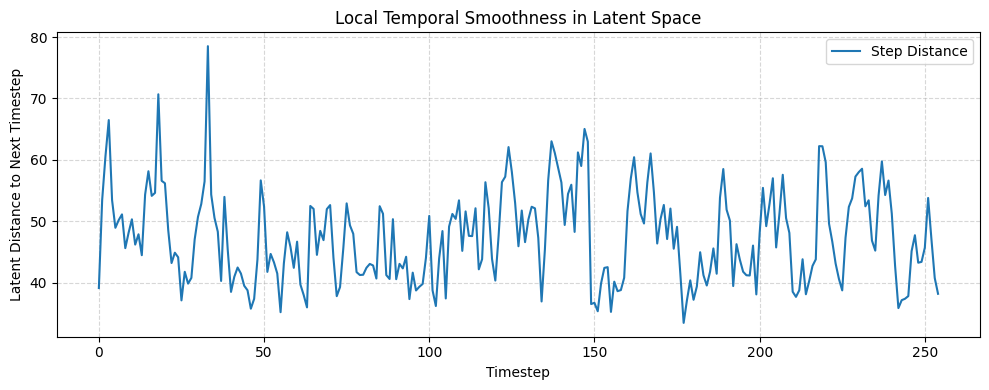}
        \caption{Autoencoder}
    \end{subfigure}
    \begin{subfigure}[b]{0.48\textwidth}
        \includegraphics[width=\textwidth]{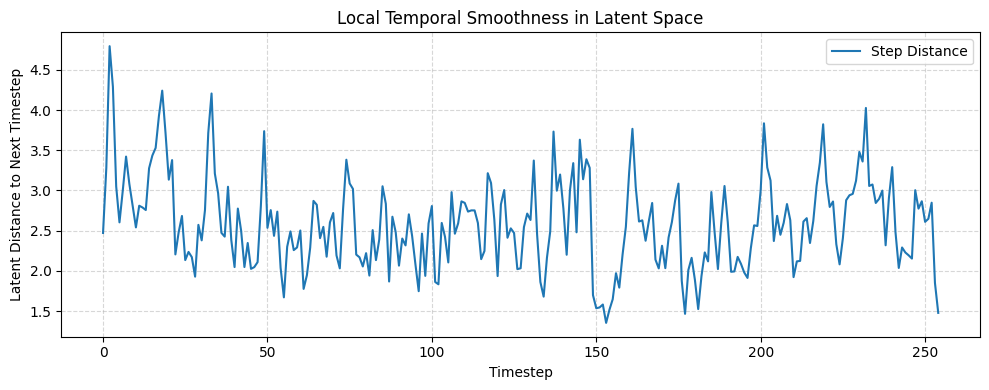}
        \caption{SPARTA - Early-Fusion HN + Cycle}
    \end{subfigure}
    \caption{Temporal Distance - Autoencoder vs SPARTA}
    \label{fig:p4_temporal_distances}
\end{figure}

\begin{figure}[H]
    \centering
    \begin{subfigure}[b]{0.48\textwidth}
        \includegraphics[width=\textwidth]{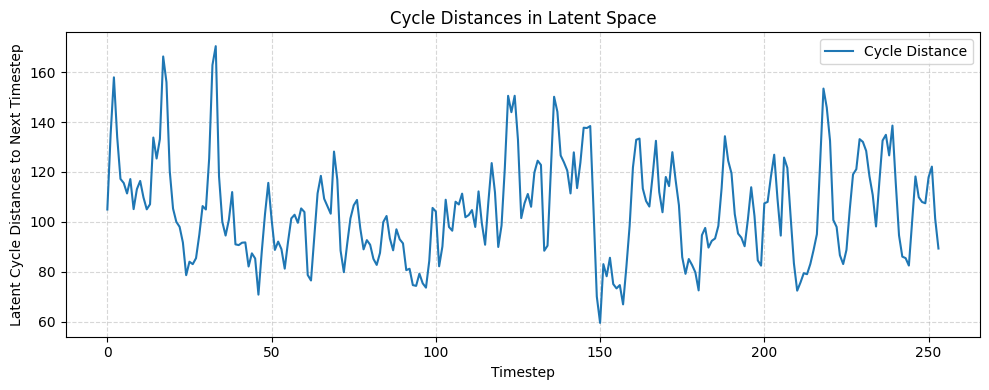}
        \caption{Autoencoder}
    \end{subfigure}
    \begin{subfigure}[b]{0.48\textwidth}
        \includegraphics[width=\textwidth]{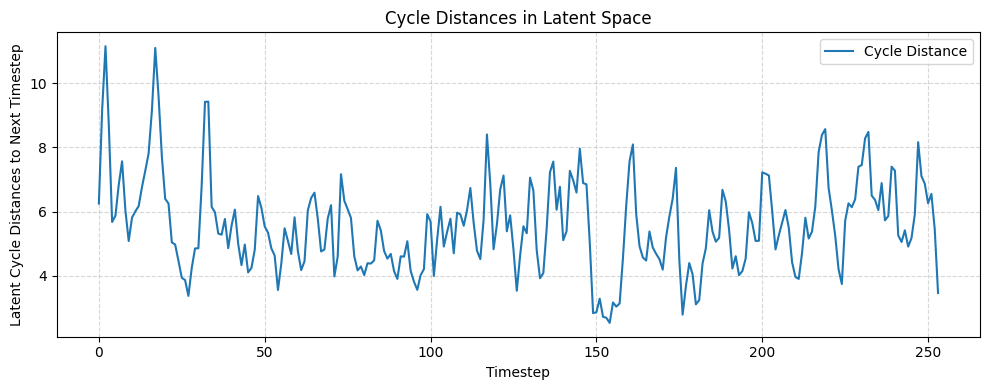}
        \caption{SPARTA - Early-Fusion HN + Cycle}
    \end{subfigure}
    \caption{Cycle Distance - Autoencoder vs SPARTA}
    \label{fig:p4_cycle_distances}
\end{figure}

We plot the temporal and cycle distances in figures \ref{fig:p4_temporal_distances} and \ref{fig:p4_cycle_distances} and additionally show the mean statistics in table \ref{table:res_5}. The table and figures show that our SPARTA model yields significantly lower temporal and cycle distances than the autoencoder, resulting in a smoother latent space. 

\begin{table}[H]
\caption{Smoothness Metrics}
\label{table:res_5}
\centering
\resizebox{\textwidth}{!}{%
\begin{tabular}{|l|c|c|}
\hline
\multicolumn{1}{|c|}{\textbf{}}           & \textbf{Mean Temporal Distance} & \textbf{Mean Cycle Distance} \\ \hline
\textbf{Autoencoder} & $47.39 \pm 7.53$                & $104.96 \pm 20.12$           \\ \hline
\textbf{SPARTA - Early-Fusion HN + Cycle}                      & $2.59 \pm 0.57$                 & $5.56 \pm 1.45$              \\ \hline
\end{tabular}
}
\end{table}

\begin{figure}[H]
    \centering
    \begin{subfigure}[b]{0.48\textwidth}
        \includegraphics[width=\textwidth]{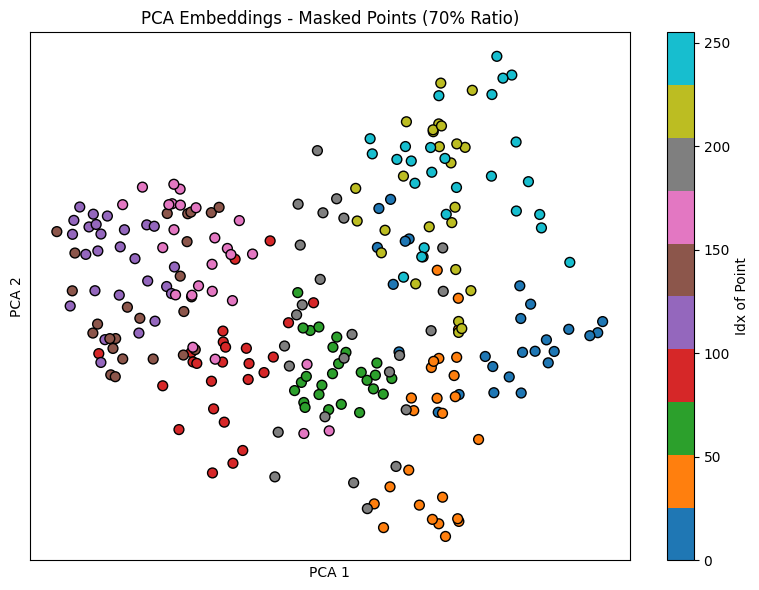}
        \caption{Autoencoder PCA}
    \end{subfigure}
    \begin{subfigure}[b]{0.48\textwidth}
        \includegraphics[width=\textwidth]{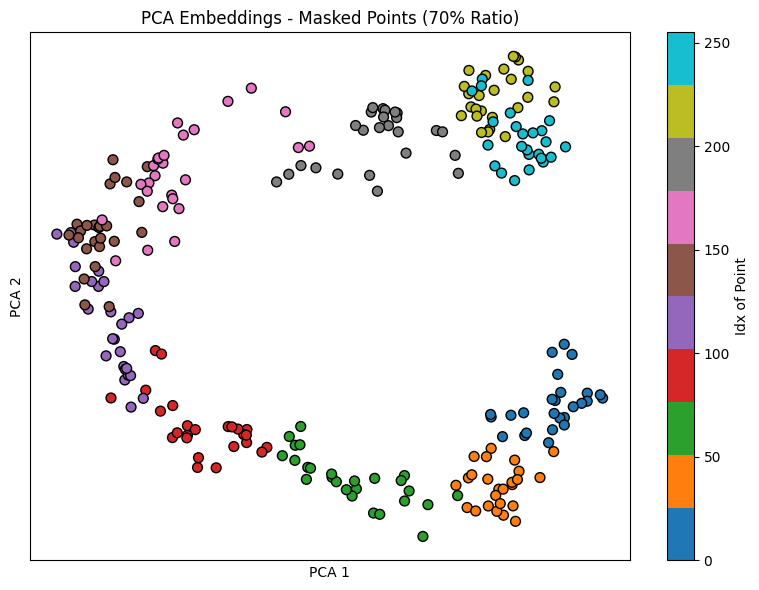}
        \caption{SPARTA - Early-Fusion HN + Cycle PCA}
    \end{subfigure}
    \caption{PCA of Latent Points Autoencoder vs SIMCLR}
    \label{fig:p4_points_2}
\end{figure}

\begin{figure}[H]
    \centering
    \begin{subfigure}[b]{0.48\textwidth}
        \includegraphics[width=\textwidth]{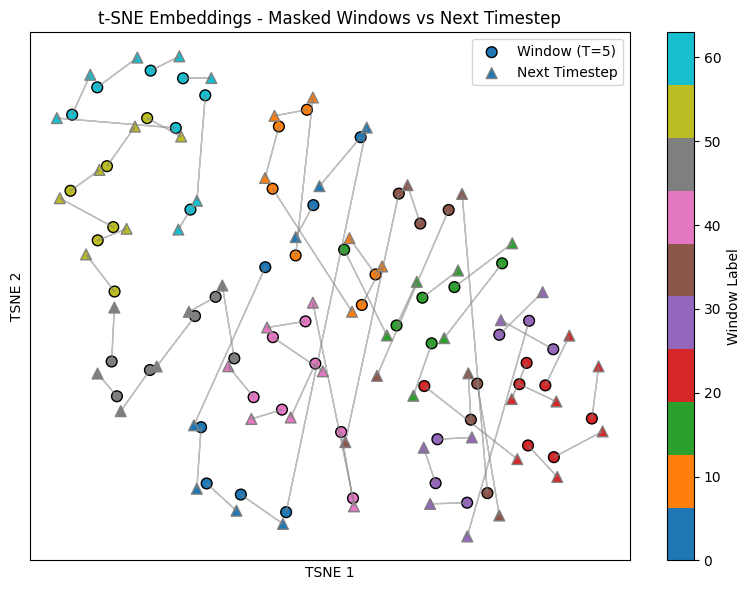}
        \caption{Autoencoder}
    \end{subfigure}
    \begin{subfigure}[b]{0.48\textwidth}
        \includegraphics[width=\textwidth]{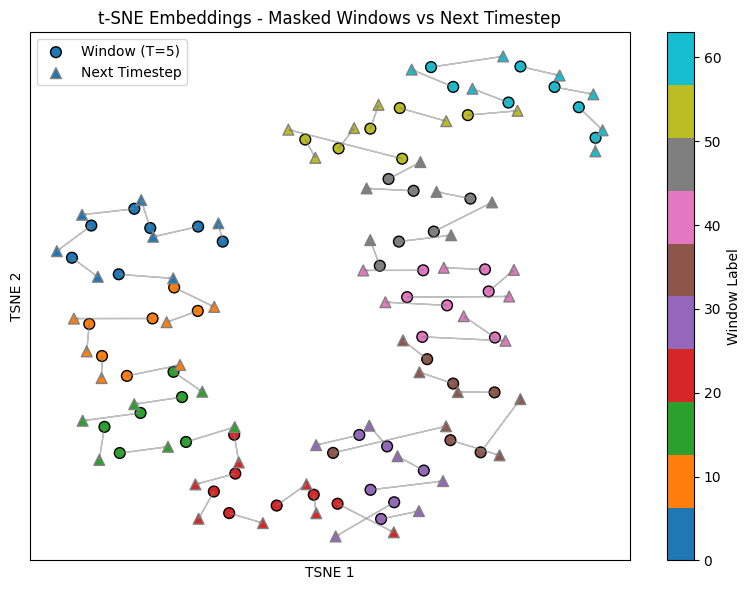}
        \caption{SPARTA - Early-Fusion HN + Cycle}
    \end{subfigure}
    \caption{Latent Trajectories – CW 5, Sampling Interval 5 Autoencoder vs SIMCLR}
    \label{fig:p4_window_points_2}
\end{figure}

We further demonstrate the benefits of contrastive learning and SPARTA by plotting latent consecutive timesteps using PCA in figure \ref{fig:p4_points_2}, as well as latent TSNE trajectories in figure \ref{fig:p4_window_points_2} for 70\% masked look-back windows of size 5, using a sampling interval of 5. Here, we show the average embedding for the look-back window and the direction of the next point to be forecasted. The plots further corroborate our findings, showing that contrastive learning produces smoother, better-structured latent points and trajectories than the autoencoder, with reduced overlap between look-back windows.

\begin{figure}[H]
    \centering
    \includegraphics[width=\textwidth]{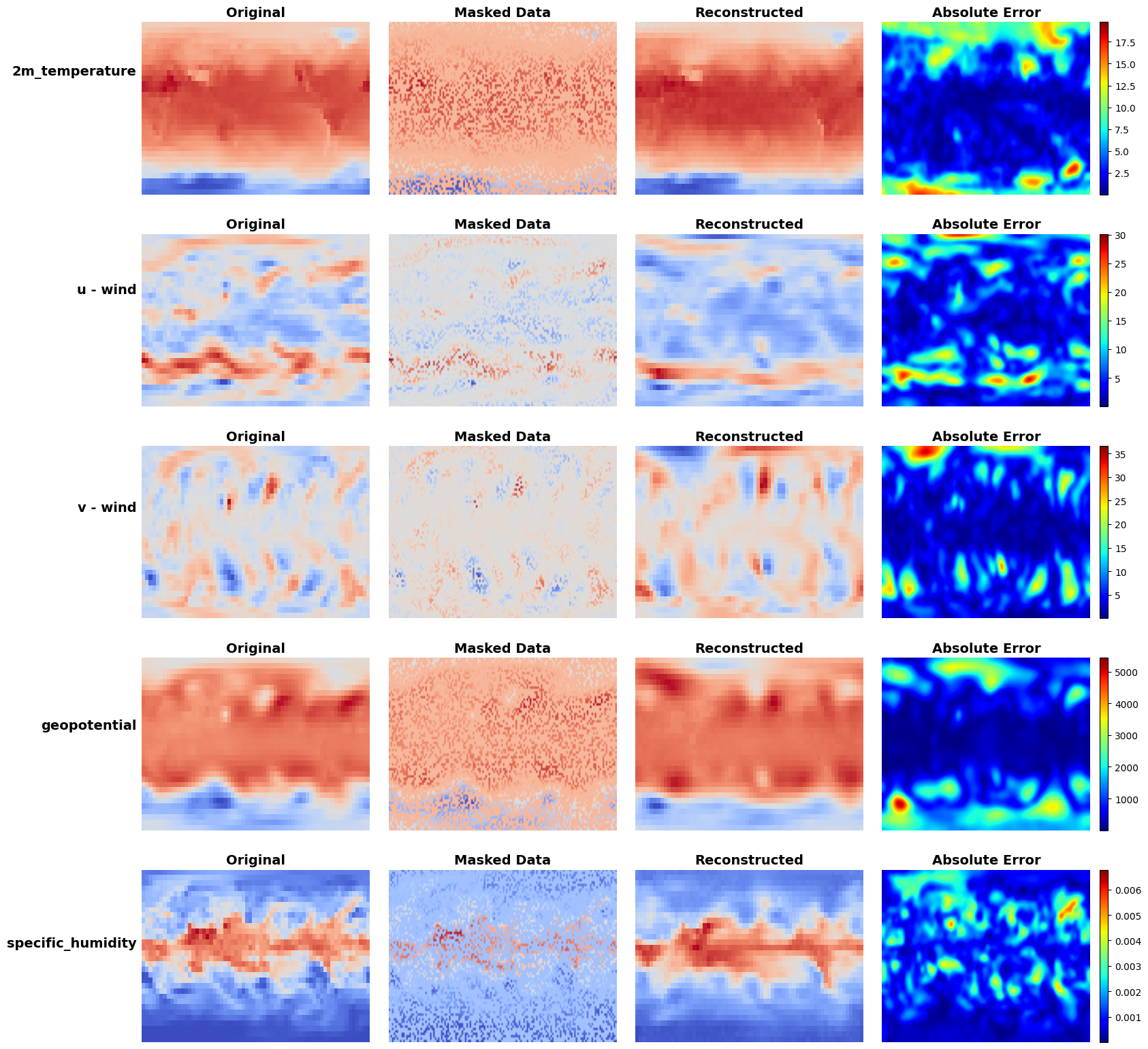}
    \caption{Forecasting Reconstruction Errors for Autoencoder - LB 30, S 1 at T=50}
    \label{fig:recon_auto}
\end{figure}

\begin{figure}[H]
    \centering
    \includegraphics[width=\textwidth]{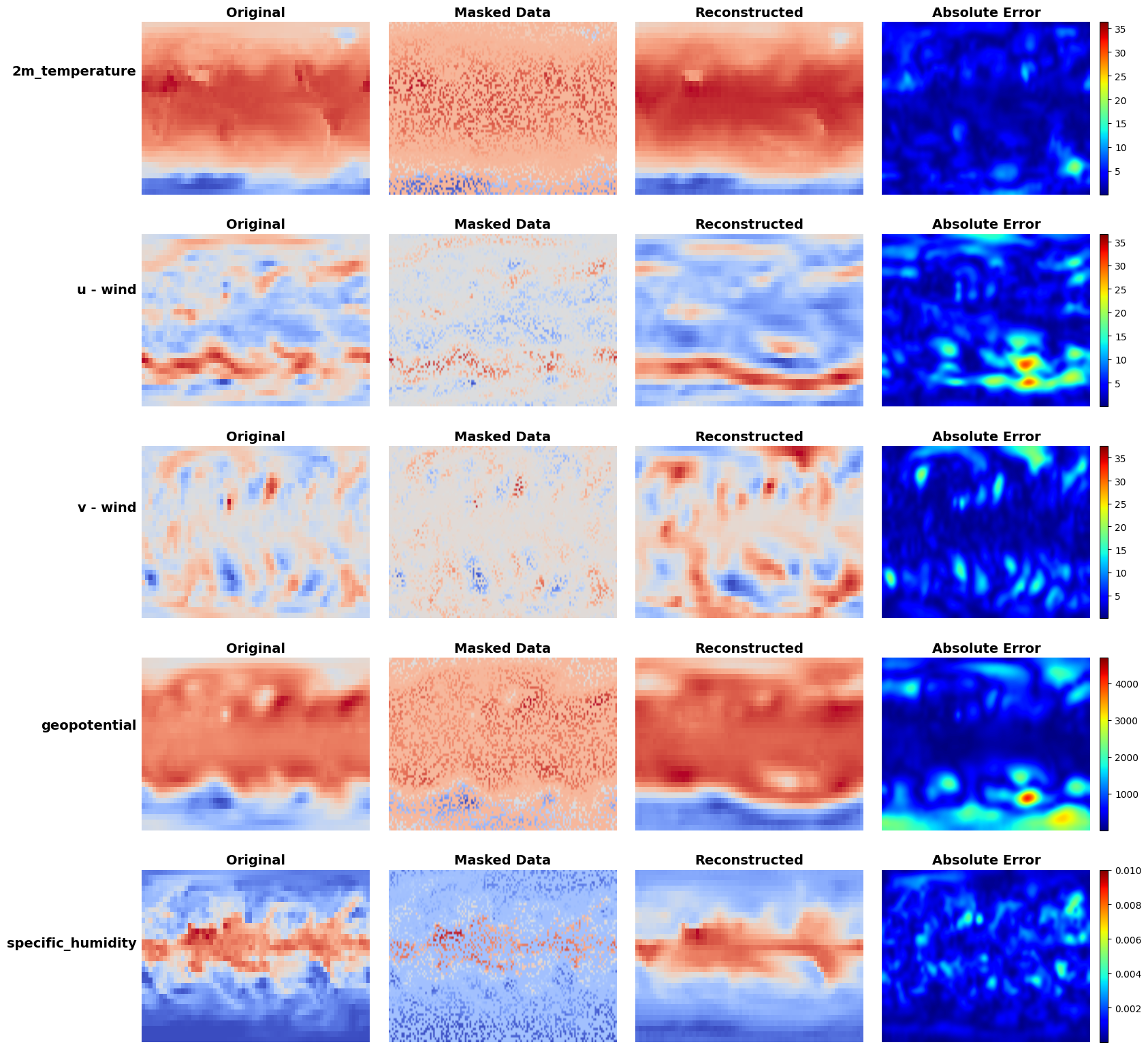}
    \caption{Forecasting Reconstruction Errors for SPARTA - Early-Fusion HN + Cycle - LB 30, S 1 at T=50}
    \label{fig:recon}
\end{figure}

A smoother latent space with better temporally structured points yields more predictable embeddings over time, making the forecasting task easier for the LSTM model and, in turn, leading to better results, as shown in the tables. Figures \ref{fig:recon_auto} and \ref{fig:recon} qualitatively show the improvement in forecasting of early fusion SPARTA compared to the autoencoder for a look-back window of 30 and a sampling interval of 1, highlighting the significant reduction in absolute errors for forecasting of a random sample of data across all modes. 

\subsubsection{Conditional Latent Diffusion and Latent Classification}

\begin{table}[H]
\caption{Conditional Latent Diffusion and Latent Classification Results}
\label{table:res_4}
\centering
\resizebox{\textwidth}{!}{%
\begin{tabular}{|l|cccc|c|}
\hline
                                              & \multicolumn{4}{c|}{{\ul \textbf{Conditional Latent Diffusion}}}                                                                                                                    & \multicolumn{1}{l|}{{\ul \textbf{Latent Classification}}} \\ \hline
\multicolumn{1}{|c|}{\textbf{}}               & \multicolumn{1}{c|}{\textbf{RRMSE (\%)}} & \multicolumn{1}{l|}{\textbf{Latent Density Score}} & \multicolumn{1}{l|}{\textbf{Realism Score}} & \multicolumn{1}{l|}{\textbf{Std Dev}} & {\ul \textbf{Test Loss}}                                  \\ \hline
\textbf{SPARTA - Early-Fusion HN + Cycle}     & \multicolumn{1}{c|}{12.19}               & \multicolumn{1}{c|}{0.92}                          & \multicolumn{1}{c|}{0.90}                   & 0.4695                                & 0.1640                                                    \\ \hline
\textbf{Autoencoder}                          & \multicolumn{1}{c|}{12.15}               & \multicolumn{1}{c|}{0.66}                          & \multicolumn{1}{c|}{1.13}                   & 0.5358                                & 0.1935                                                    \\ \hline
\end{tabular}
}
\end{table}

Table \ref{table:res_4} shows the results for the latent diffusion task. Overall, our SPARTA model achieves an RRMSE comparable to that of the autoencoder. However, incorporating contrastive learning yields a better latent space for diffusion, as evidenced by a lower standard deviation of generated latent samples and a higher latent density score. \\

Contrastive learning focuses on creating embeddings that preserve shared information across views, while ignoring the augmentations applied to create them. This leads to the applied augmentations (e.g. noise, jitter) being ignored in the latent space. This produces a latent space that highlights the data's semantic structure. Small perturbations do not move the latent code far from that of the clean data. Furthermore, Wang et al \cite{p4_4} show that contrastive learning produces latent codes that are aligned and uniformly distributed on a hypersphere. Uniformity means that latent codes are evenly distributed, so adding noise will not move the latent vector far from the training distribution of encoded latent vectors. Alignment means that adding noise does not move the latent vector far from its clean counterpart. In addition, the introduction of contrastive loss ensures greater separation between negative samples, meaning that perturbing a latent vector slightly with noise will not move the vector into the realm of a vector of a different data sample.  \\

For these reasons, the denoised latent codes will be closer to clean samples than the autoencoder's latent codes. This is confirmed in table \ref{table:res_4} by the higher latent density score of SPARTA compared to the autoencoder. As mentioned in section \ref{sec:model_eval}, the latent density score measures the Gaussian kernelised Euclidean distance between the generated sample and the ground truth. A higher score indicates that the generated samples are closer to the ground-truth clean samples, confirming the findings of Wang et al \cite{p4_4}, which show that contrastive learning creates a more uniform and aligned latent space. In addition, the reduced standard deviation in the generation of our samples, as shown in table \ref{table:res_4}, shows that SPARTA creates a more structured latent space, where the noised latent sample has a clearer and less ambiguous path back to the clean latent sample. Overall, our results in table \ref{table:res_4} show that SPARTA creates a more robust latent space compared to the autoencoder.\\

Table \ref{table:res_4} also shows that our SPARTA model creates a latent space that is better primed for latent classification than the autoencoder benchmark. The discriminative objective of contrastive learning enforces an explicit separation between negative and positive samples, thereby increasing the linear separability of the latent space \cite{p4_2}, thereby improving classification performance. \\

\begin{figure}[H]
    \centering
    \includegraphics[width=\textwidth]{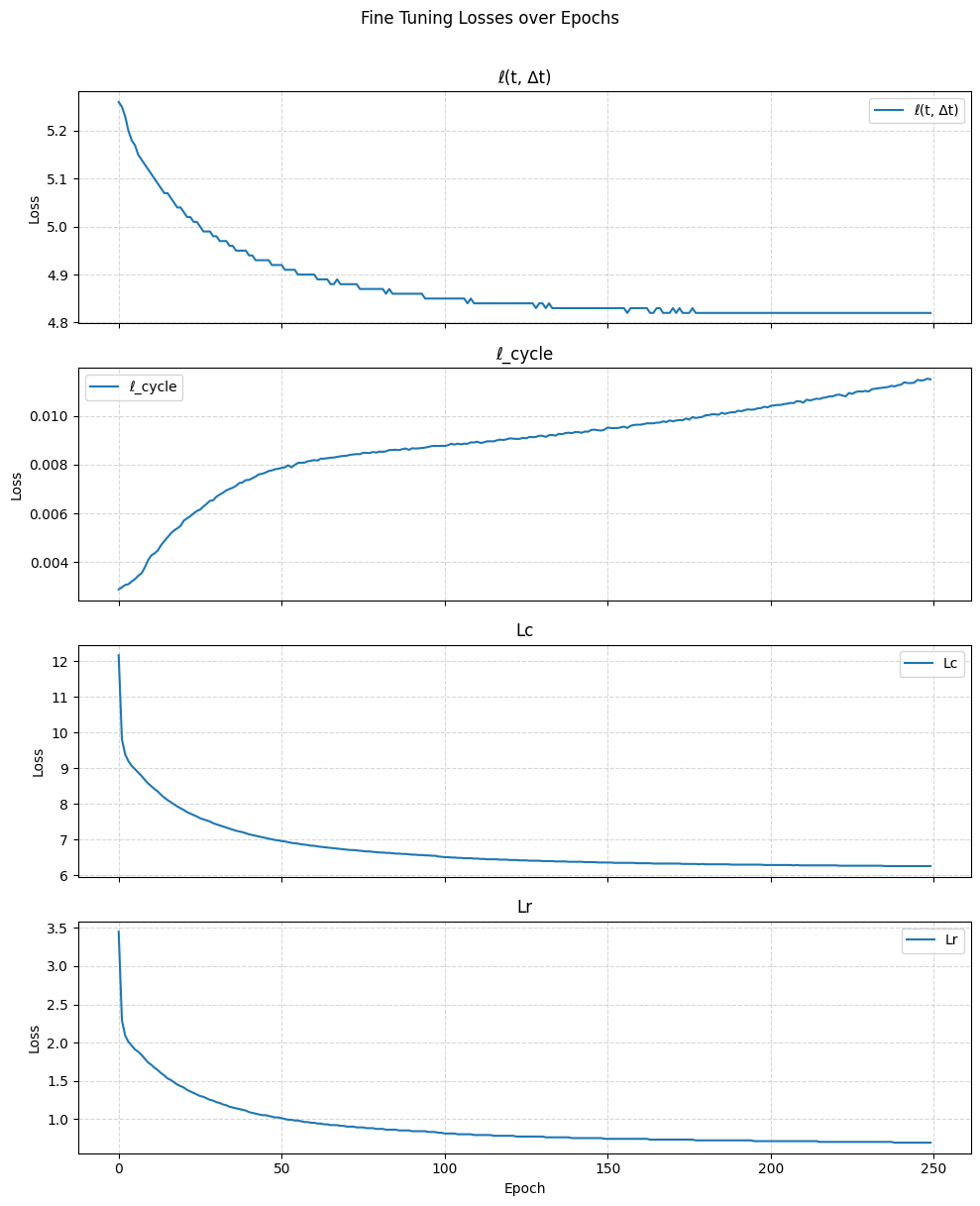}
    \caption{Training Losses - Stage 2}
    \label{fig:fine_tune}
\end{figure}

\begin{figure}[H]
    \centering
    \includegraphics[width=0.7\textwidth]{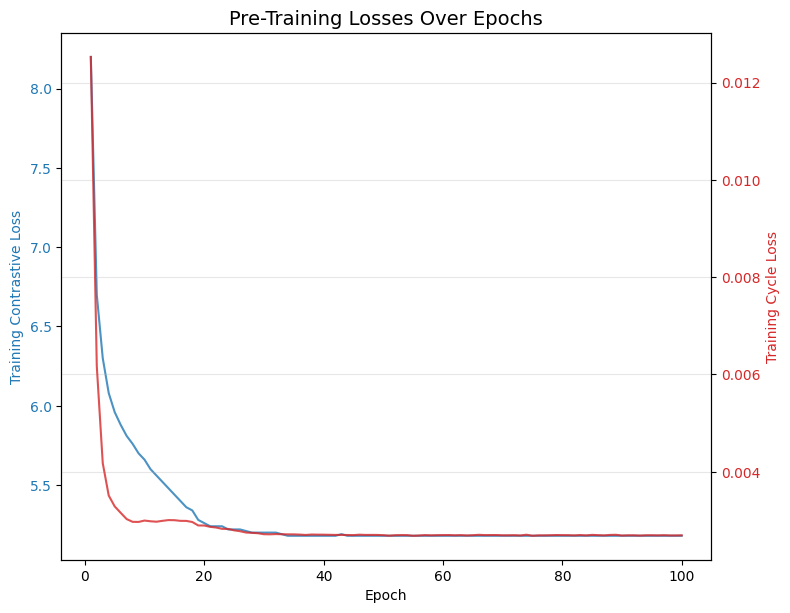}
    \caption{Training Losses - Stage 1}
    \label{fig:pre_train}
\end{figure}

Finally, we show the losses for training stage two of SPARTA in figure \ref{fig:fine_tune}, as explained in section \ref{sec:train}. This highlights a stable and convergent training process for SPARTA. The individual and total contrastive losses reach a plateau, indicating that our model has converged and that our samples are better separated in the latent space. Whilst there is a slight increase in cycle loss, we theorise that this is due to the alpha-decay term applied to the contrastive loss relative to the reconstruction loss. This is supported by figure \ref{fig:pre_train}, which highlights the losses for stage one of the training process. We can see from this figure that without the decaying alpha term, the cycle loss decreases and converges. Lastly, we observe a monotonic decrease in the reconstruction loss during stage two, indicating that the decoder is improving its ability to reconstruct the multi-modal embeddings. Similar to the contrastive losses, it also converges and reaches a plateau. As we observe both contrastive and reconstruction losses decreasing and plateauing, we conclude that our joint training procedure is successful and that SPARTA can couple contrastive and reconstruction learning.

\subsection{Ablation Test}

\subsubsection{Forecasting}

\begin{table}[H]
\caption{Autoregressive Results - 100 Steps LB 30, S 1}
\label{table:res_1_ab}
\centering
\resizebox{\textwidth}{!}{%
\begin{tabular}{|l|cccc|}
\hline
                                              & \multicolumn{4}{c|}{{\ul \textbf{Look-Back Window 30, Sampling Interval 1}}}                                                                                                                                \\ \hline
\multicolumn{1}{|c|}{\textbf{}}               & \multicolumn{1}{c|}{\textbf{RRMSE (\%) T=25}} & \multicolumn{1}{l|}{\textbf{RRMSE (\%) T=50}} & \multicolumn{1}{l|}{\textbf{RRMSE (\%) T=75}} & \multicolumn{1}{l|}{\textbf{RRMSE (\%) T=100}} \\ \hline
\textbf{SPARTA - Early-Fusion}                & \multicolumn{1}{c|}{22.92}                    & \multicolumn{1}{c|}{23.68}                    & \multicolumn{1}{c|}{24.17}                    & 24.50                                          \\ \hline
\textbf{SPARTA - Early-Fusion Cycle}          & \multicolumn{1}{c|}{21.63}                    & \multicolumn{1}{c|}{22.06}                    & \multicolumn{1}{c|}{22.42}                    & 22.92                                          \\ \hline
\textbf{SPARTA - Early-Fusion HN}             & \multicolumn{1}{c|}{18.44}                    & \multicolumn{1}{c|}{20.17}                    & \multicolumn{1}{c|}{20.39}                    & 20.67                                          \\ \hline
\textbf{SPARTA - Early-Fusion HN + Cycle}     & \multicolumn{1}{c|}{\textbf{15.71}}           & \multicolumn{1}{c|}{\textbf{16.45}}           & \multicolumn{1}{c|}{\textbf{16.84}}           & \textbf{17.04}                                 \\ \hline
\end{tabular}
}
\end{table}

\begin{table}[H]
\caption{Autoregressive Results - LB 5, S 5}
\label{table:res_2_ab}
\centering
\resizebox{\textwidth}{!}{%
\begin{tabular}{|l|cccc|}
\hline
                                              & \multicolumn{4}{c|}{{\ul \textbf{Look-Back Window 5, Sampling Interval 5}}}                                                                                                                                 \\ \hline
\multicolumn{1}{|c|}{\textbf{}}               & \multicolumn{1}{c|}{\textbf{RRMSE (\%) T=25}} & \multicolumn{1}{l|}{\textbf{RRMSE (\%) T=50}} & \multicolumn{1}{l|}{\textbf{RRMSE (\%) T=75}} & \multicolumn{1}{l|}{\textbf{RRMSE (\%) T=100}} \\ \hline
\textbf{SPARTA - Early-Fusion}                & \multicolumn{1}{c|}{18.87}                    & \multicolumn{1}{c|}{20.72}                    & \multicolumn{1}{c|}{21.57}                    & 22.02                                          \\ \hline
\textbf{SPARTA - Early-Fusion Cycle}          & \multicolumn{1}{c|}{22.58}                    & \multicolumn{1}{c|}{25.01}                    & \multicolumn{1}{c|}{24.95}                    & 25.16                                          \\ \hline
\textbf{SPARTA - Early-Fusion HN}             & \multicolumn{1}{c|}{17.48}                    & \multicolumn{1}{c|}{17.04}                    & \multicolumn{1}{c|}{16.61}                    & 16.57                                          \\ \hline
\textbf{SPARTA - Early-Fusion HN + Cycle}     & \multicolumn{1}{c|}{14.89}                    & \multicolumn{1}{c|}{17.30}                    & \multicolumn{1}{c|}{17.74}                    & 17.71                                          \\ \hline
\end{tabular}
}
\end{table}

\begin{table}[H]
\caption{Autoregressive Results - 100 Steps LB 5, S 10}
\label{table:res_3_ab}
\centering
\resizebox{\textwidth}{!}{%
\begin{tabular}{|l|cccc|}
\hline
                                              & \multicolumn{4}{c|}{{\ul \textbf{Look-Back Window 5, Sampling Interval 10}}}                                                                                                                                \\ \hline
\multicolumn{1}{|c|}{\textbf{}}               & \multicolumn{1}{c|}{\textbf{RRMSE (\%) T=25}} & \multicolumn{1}{l|}{\textbf{RRMSE (\%) T=50}} & \multicolumn{1}{l|}{\textbf{RRMSE (\%) T=75}} & \multicolumn{1}{l|}{\textbf{RRMSE (\%) T=100}} \\ \hline
\textbf{SPARTA - Early-Fusion}                & \multicolumn{1}{c|}{18.91}                    & \multicolumn{1}{c|}{19.25}                    & \multicolumn{1}{c|}{19.03}                    & 19.26                                          \\ \hline
\textbf{SPARTA - Early-Fusion Cycle}          & \multicolumn{1}{c|}{19.63}                    & \multicolumn{1}{c|}{20.45}                    & \multicolumn{1}{c|}{20.70}                    & 21.54                                          \\ \hline
\textbf{SPARTA - Early-Fusion HN}             & \multicolumn{1}{c|}{18.64}                    & \multicolumn{1}{c|}{19.46}                    & \multicolumn{1}{c|}{19.51}                    & 19.69                                          \\ \hline
\textbf{SPARTA - Early-Fusion HN + Cycle}     & \multicolumn{1}{c|}{13.62}                    & \multicolumn{1}{c|}{14.11}                    & \multicolumn{1}{c|}{14.75}                    & 15.03                                          \\ \hline
\end{tabular}
}
\end{table}

\subsubsection{Conditional Latent Diffusion and Latent Classification}

\begin{table}[H]
\caption{Conditional Latent Diffusion and Latent Classification Results}
\label{table:res_4_ab}
\centering
\resizebox{\textwidth}{!}{%
\begin{tabular}{|l|cccc|c|}
\hline
                                              & \multicolumn{4}{c|}{{\ul \textbf{Conditional Latent Diffusion}}}                                                                                                                    & \multicolumn{1}{l|}{{\ul \textbf{Latent Classification}}} \\ \hline
\multicolumn{1}{|c|}{\textbf{}}               & \multicolumn{1}{c|}{\textbf{RRMSE (\%)}} & \multicolumn{1}{l|}{\textbf{Latent Density Score}} & \multicolumn{1}{l|}{\textbf{Realism Score}} & \multicolumn{1}{l|}{\textbf{Std Dev}} & {\ul \textbf{Test Loss}}                                  \\ \hline
\textbf{SPARTA - Early-Fusion}                & \multicolumn{1}{c|}{\textbf{11.43}}      & \multicolumn{1}{c|}{\textbf{0.91}}                 & \multicolumn{1}{c|}{\textbf{1.00}}          & \textbf{0.4265}                       & \textbf{0.1454}                                                    \\ \hline
\textbf{SPARTA - Early-Fusion Cycle}          & \multicolumn{1}{c|}{11.65}               & \multicolumn{1}{c|}{0.92}                          & \multicolumn{1}{c|}{0.98}                   & 0.4456                                & 0.1620                                                    \\ \hline
\textbf{SPARTA - Early-Fusion HN}             & \multicolumn{1}{c|}{11.64}               & \multicolumn{1}{c|}{0.91}                          & \multicolumn{1}{c|}{1.02}                   & 0.4551                                & 0.1597                                                    \\ \hline
\textbf{SPARTA - Early-Fusion HN + Cycle}     & \multicolumn{1}{c|}{12.19}               & \multicolumn{1}{c|}{0.92}                          & \multicolumn{1}{c|}{0.90}                   & 0.4695                                & 0.1640                                                    \\ \hline
\end{tabular}
}
\end{table}

To highlight the impact of our novel contributions, we perform an ablation test by removing each contribution to assess its effect on performance. \\

Interestingly, we find that the cycle consistency loss leads to a smoother latent space, as highlighted in table \ref{table:res_smooth}. We observe an improvement in forecasting only with a sampling interval of 1 and a look-back window of 30, as highlighted by table \ref{table:res_1_ab}. The inclusion of the hard negative sampling scheme has a greater impact, as shown in tables \ref{table:res_1_ab} and \ref{table:res_2_ab}, improving performance across the first two forecasting experiments. Table \ref{table:res_smooth} shows that introducing this sampling scheme actually makes the latent space less smooth. Whilst we would normally expect this to lead to a decrease in forecasting performance, the additional challenge of separating hard and soft negatives results in higher-quality embeddings. We theorise that this change in embedding quality thereby improves performance to a greater degree than the decrease in performance caused by the less smooth latent space. \\

\begin{figure}[H]
    \centering
    \includegraphics[width=\textwidth]{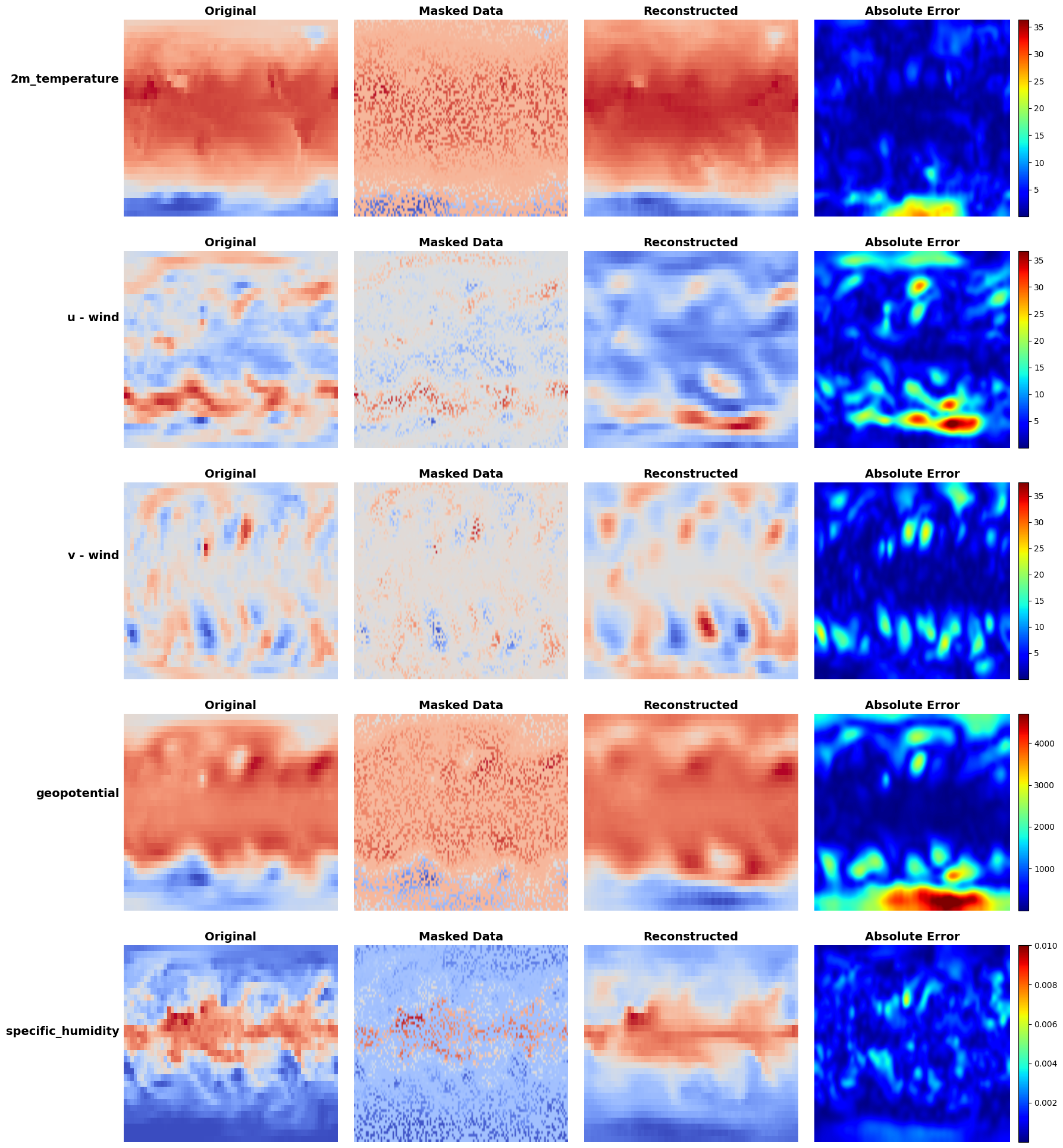}
    \caption{Forecasting Errors SPARTA - Early-Fusion - LB 30, S 1 at T=50}
    \label{fig:recon_abs}
\end{figure}

When combining both novel changes, tables \ref{table:res_1_ab}, \ref{table:res_2_ab}, and \ref{table:res_3_ab} show improvements across all forecasting experiments, as further explained by the increased smoothness highlighted in table \ref{table:res_smooth}. We qualitatively show the forecasting improvement of our novel changes in figures \ref{fig:recon} and \ref{fig:recon_abs} for a look-back window of 30 and a sampling interval of 1. \\

However, we note that introducing these changes negatively affects diffusion and classification results, as shown in table \ref{table:res_4_ab}, suggesting a trade-off in downstream task performance when encouraging a smoother latent space. In conclusion, we demonstrate that combining these two approaches is necessary to produce a smoother latent space and outperform the SimCLR benchmark on the forecasting task. 

\begin{table}[H]
\caption{Smoothness Metrics - Ablation Test}
\label{table:res_smooth}
\centering
\resizebox{\textwidth}{!}{%
\begin{tabular}{|l|c|c|}
\hline
\multicolumn{1}{|c|}{\textbf{}}           & \textbf{Mean Temporal Distance} & \textbf{Mean Cycle Distance} \\ \hline
\textbf{SPARTA - Early-Fusion}            & $4.25 \pm 0.86$                 & $8.77 \pm 2.09$              \\ \hline
\textbf{SPARTA - Early-Fusion Cycle}      & $3.27 \pm 0.70$                 & $6.85 \pm 1.72$              \\ \hline
\textbf{SPARTA - Early-Fusion HN}         & $4.36 \pm 0.92$                 & $9.22 \pm 2.35$              \\ \hline
\textbf{SPARTA - Early-Fusion HN + Cycle} & $2.59 \pm 0.57$                 & $5.56 \pm 1.45$              \\ \hline
\end{tabular}
}
\end{table}

\subsection{Late-Fusion}
\vspace{3mm}

\begin{table}[H]
\caption{Autoregressive Results - 100 Steps LB 30, S 1}
\label{table:res_1_late}
\centering
\resizebox{\textwidth}{!}{%
\begin{tabular}{|l|cccc|}
\hline
                                              & \multicolumn{4}{c|}{{\ul \textbf{Look-Back Window 30, Sampling Interval 1}}}                                                                                                                                \\ \hline
\multicolumn{1}{|c|}{\textbf{}}               & \multicolumn{1}{c|}{\textbf{RRMSE (\%) T=25}} & \multicolumn{1}{l|}{\textbf{RRMSE (\%) T=50}} & \multicolumn{1}{l|}{\textbf{RRMSE (\%) T=75}} & \multicolumn{1}{l|}{\textbf{RRMSE (\%) T=100}} \\ \hline
\textbf{SPARTA - Early-Fusion HN + Cycle}     & \multicolumn{1}{c|}{\textbf{15.71}}           & \multicolumn{1}{c|}{\textbf{16.45}}           & \multicolumn{1}{c|}{\textbf{16.84}}           & \textbf{17.04}                                 \\ \hline
\textbf{SPARTA - Late-Fusion, Self-Attention} & \multicolumn{1}{c|}{16.97}                    & \multicolumn{1}{c|}{18.26}                    & \multicolumn{1}{c|}{19.00}                    & 19.54                                          \\ \hline
\textbf{SPARTA - Late-Fusion, GNN}            & \multicolumn{1}{c|}{16.36}                    & \multicolumn{1}{c|}{18.78}                    & \multicolumn{1}{c|}{18.94}                    & 18.76                                          \\ \hline
\end{tabular}
}
\end{table}

\begin{table}[H]
\caption{Autoregressive Results - 100 Steps LB 5, S 5}
\label{table:res_2_late}
\centering
\resizebox{\textwidth}{!}{%
\begin{tabular}{|l|cccc|}
\hline
                                              & \multicolumn{4}{c|}{{\ul \textbf{Look-Back Window 5, Sampling Interval 5}}}                                                                                                                                 \\ \hline
\multicolumn{1}{|c|}{\textbf{}}               & \multicolumn{1}{c|}{\textbf{RRMSE (\%) T=25}} & \multicolumn{1}{l|}{\textbf{RRMSE (\%) T=50}} & \multicolumn{1}{l|}{\textbf{RRMSE (\%) T=75}} & \multicolumn{1}{l|}{\textbf{RRMSE (\%) T=100}} \\ \hline
\textbf{SPARTA - Early-Fusion HN + Cycle}     & \multicolumn{1}{c|}{14.89}                    & \multicolumn{1}{c|}{17.30}                    & \multicolumn{1}{c|}{17.74}                    & 17.71                                          \\ \hline
\textbf{SPARTA - Late-Fusion, Self-Attention} & \multicolumn{1}{c|}{14.02}                    & \multicolumn{1}{c|}{15.60}                    & \multicolumn{1}{c|}{15.99}                    & 15.93                                          \\ \hline
\textbf{SPARTA - Late-Fusion, GNN}            & \multicolumn{1}{c|}{\textbf{11.41}}           & \multicolumn{1}{c|}{\textbf{11.98}}           & \multicolumn{1}{c|}{\textbf{12.56}}           & \textbf{12.85}                                 \\ \hline
\end{tabular}
}
\end{table}

\begin{table}[H]
\caption{Autoregressive Results - 100 Steps LB 5, S 10}
\label{table:res_3_late}
\centering
\resizebox{\textwidth}{!}{%
\begin{tabular}{|l|cccc|}
\hline
                                              & \multicolumn{4}{c|}{{\ul \textbf{Look-Back Window 5, Sampling Interval 10}}}                                                                                                                                \\ \hline
\multicolumn{1}{|c|}{\textbf{}}               & \multicolumn{1}{c|}{\textbf{RRMSE (\%) T=25}} & \multicolumn{1}{l|}{\textbf{RRMSE (\%) T=50}} & \multicolumn{1}{l|}{\textbf{RRMSE (\%) T=75}} & \multicolumn{1}{l|}{\textbf{RRMSE (\%) T=100}} \\ \hline
\textbf{SPARTA - Early-Fusion HN + Cycle}     & \multicolumn{1}{c|}{13.62}                    & \multicolumn{1}{c|}{14.11}                    & \multicolumn{1}{c|}{14.75}                    & 15.03                                          \\ \hline
\textbf{SPARTA - Late-Fusion, Self-Attention} & \multicolumn{1}{c|}{12.99}                    & \multicolumn{1}{c|}{13.20}                    & \multicolumn{1}{c|}{13.16}                    & 13.36                                          \\ \hline
\textbf{SPARTA - Late-Fusion, GNN}            & \multicolumn{1}{c|}{\textbf{12.11}}           & \multicolumn{1}{c|}{\textbf{12.68}}           & \multicolumn{1}{c|}{\textbf{12.63}}           & \textbf{12.77}                                 \\ \hline
\end{tabular}
}
\end{table}

\subsubsection{Conditional Latent Diffusion and Latent Classification}

\begin{table}[H]
\caption{Conditional Latent Diffusion and Latent Classification Results}
\label{table:res_4_late}
\centering
\resizebox{\textwidth}{!}{%
\begin{tabular}{|l|cccc|c|}
\hline
                                              & \multicolumn{4}{c|}{{\ul \textbf{Conditional Latent Diffusion}}}                                                                                                                    & \multicolumn{1}{l|}{{\ul \textbf{Latent Classification}}} \\ \hline
\multicolumn{1}{|c|}{\textbf{}}               & \multicolumn{1}{c|}{\textbf{RRMSE (\%)}} & \multicolumn{1}{l|}{\textbf{Latent Density Score}} & \multicolumn{1}{l|}{\textbf{Realism Score}} & \multicolumn{1}{l|}{\textbf{Std Dev}} & {\ul \textbf{Test Loss}}                                  \\ \hline
\textbf{SPARTA - Early-Fusion HN + Cycle}     & \multicolumn{1}{c|}{12.19}               & \multicolumn{1}{c|}{0.92}                          & \multicolumn{1}{c|}{0.90}                   & 0.4695                                & 0.1640                                                    \\ \hline
\textbf{SPARTA - Late-Fusion, Self-Attention} & \multicolumn{1}{c|}{12.47}               & \multicolumn{1}{c|}{0.95}                          & \multicolumn{1}{c|}{0.91}                   & 0.4352                                & \textbf{0.1264}                                           \\ \hline
\textbf{SPARTA - Late-Fusion, GNN}            & \multicolumn{1}{c|}{12.23}               & \multicolumn{1}{c|}{0.96}                          & \multicolumn{1}{c|}{0.90}                   & 0.4378                                & 0.1412                                                    \\ \hline
\end{tabular}
}
\end{table}

We contrast our early-fusion approach with the two late-fusion approaches mentioned in section \ref{sec:late_fusion}. In both late-fusion methods, we utilise both the hard negative sampling scheme and the cycle consistency loss. The results ultimately show that the self-attention method offers improved latent classification results, as highlighted by table \ref{table:res_4_late}, whereas the graph neural network approach leads to better forecasting results, which are shown in tables \ref{table:res_1_late}, \ref{table:res_2_late}, and \ref{table:res_3_late}. \\

\begin{figure}[H]
    \centering
    \begin{subfigure}[b]{0.48\textwidth}
        \includegraphics[width=\textwidth]{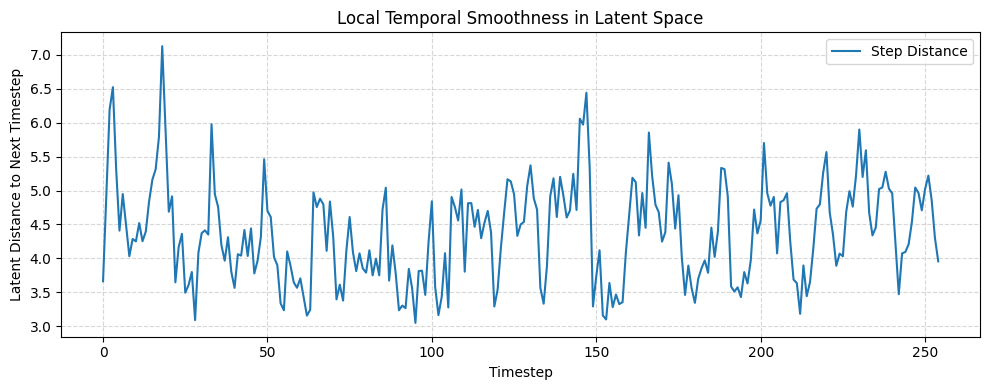}
        \caption{SPARTA - Late-Fusion, Self-Attention}
    \end{subfigure}
    \begin{subfigure}[b]{0.48\textwidth}
        \includegraphics[width=\textwidth]{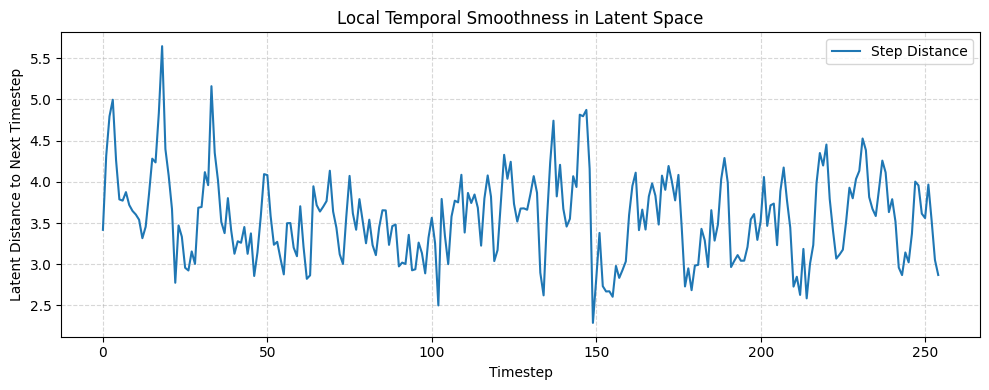}
        \caption{SPARTA - Late-Fusion, GNN}
    \end{subfigure}
    \caption{Temporal Distance - SPARTA - Late Fusion  - Self Attention vs GNN}
    \label{fig:fusion_temporal_distances}
\end{figure}

\begin{figure}[H]
    \centering
    \begin{subfigure}[b]{0.48\textwidth}
        \includegraphics[width=\textwidth]{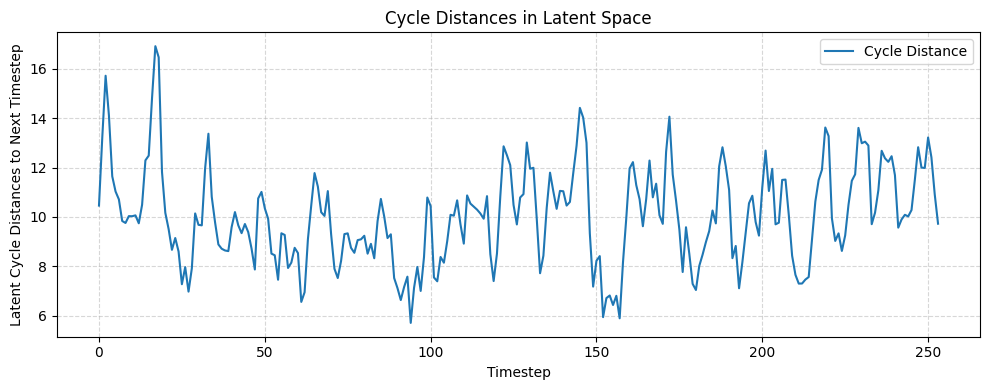}
        \caption{SPARTA - Late-Fusion, Self-Attention}
    \end{subfigure}
    \begin{subfigure}[b]{0.48\textwidth}
        \includegraphics[width=\textwidth]{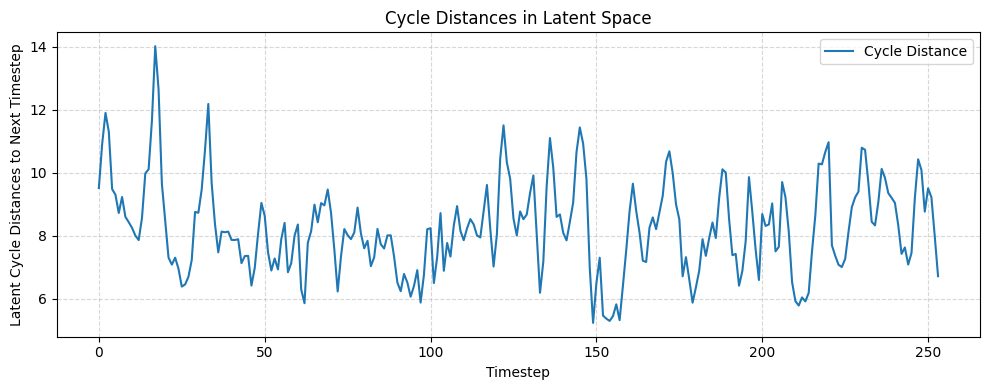}
        \caption{SPARTA - Late-Fusion, GNN}
    \end{subfigure}
    \caption{Cycle Distance - SPARTA - Late Fusion  - Self Attention vs GNN}
    \label{fig:fusion_cycle_distances}
\end{figure}

Fundamentally, self-attention does not allow for prior knowledge to be inserted. In contrast, our GNN approach offers an architecture that uniquely enables the incorporation of domain-specific knowledge via the adjacency matrix. This key distinction enhances the robustness of our model, making it a powerful alternative for forecasting. For each data sample, self-attention enables each mode to interact with one another. This allows for the creation of more flexible and richer embeddings. The GNN approach, on the other hand, adds regularisation by enforcing a prior on which modes can interact. This limits the flexibility of the embeddings but reduces temporal deviation, creating a better-structured, smoother latent space. The latent classification task is not concerned with how the embeddings evolve, but rather with identifying the most discriminative embedding for each data point. Forecasting benefits from a smoother latent space, where the embeddings evolve more predictably over time. This is evident in the reduced temporal and cycle distances, as shown in figures \ref{fig:fusion_temporal_distances} and \ref{fig:fusion_cycle_distances} and table \ref{table:p6_smooth}, which highlight a smoother, more suitable latent space for forecasting. 

\begin{table}[H]
\caption{Smoothness Metrics - Self Attention vs GNN}
\label{table:p6_smooth}
\centering
\resizebox{\textwidth}{!}{%
\begin{tabular}{|l|c|c|c|}
\hline
\multicolumn{1}{|c|}{\textbf{}}               & \textbf{Mean Temporal Distance} & \textbf{Mean Cycle Distance}  \\ \hline
\textbf{SPARTA - Late-Fusion, Self-Attention} & $4.37 \pm 0.73$                 & $10.02 \pm 2.11$               \\ \hline
\textbf{SPARTA - Late-Fusion, GNN} & $3.57 \pm 0.52$                 & $8.24 \pm 1.45$              \\ \hline
\end{tabular}
}
\end{table}

\section{Conclusion}
\vspace{3mm}

Overall, this work has extensively explored contrastive learning for ERA5 data, extending the current literature by developing an end-to-end, early-fusion contrastive learning model that is compatible with sparse data. We demonstrated that this model surpasses a simple autoencoder on the downstream forecasting task by leveraging a novel hard-negative sampling approach and a cycle-consistency loss, achieving up to a 32\% improvement in forecasting performance. \\

We further extended our evaluation, showing that contrastive learning yielded a latent space that outperformed the autoencoder across two additional diverse tasks: latent classification and latent diffusion. Our best contrastive learning model achieved up to a 23\% reduction in the standard deviation of latent diffusion and a 28\% improvement in the latent classification loss. \\

We further expanded on the current literature by proposing a novel GNN approach for multimodal fusion and demonstrating that regularisation with prior knowledge yields a smoother latent space, leading to performance improvements in forecasting compared to the self-attention fusion method.

\section{Future Work}
\vspace{3mm}

This work has expanded the analysis of contrastive learning on ERA5 data. However, there remains scope for further iteration; specifically, the following ideas could be incorporated for future work.

\begin{enumerate}
    \item \textbf{Higher Resolution Data}: Our methods only evaluated the models on the smallest resolution of ERA5 data: 64x32. This limits the usability of our solutions, preventing them from being used in tasks that require access to higher-resolution data. This was a practical consideration, as training on higher-resolution data, especially across all coordinates, requires significantly more computation. Future work, with access to additional computational resources, could train our contrastive learning models on higher-resolution ERA5 data to assess their scalability and performance at a finer scale.
    \item \textbf{Incorporating More Modalities and Varying Resolutions}: Our methods only used five modes of data. We therefore cannot assess how well our solution would perform if additional modalities were incorporated. Moreover, we assumed that each mode is collected at the same resolution, which may not always be valid. Future work could therefore expand our final model to include more modes at varying resolutions. The introduction of mixture-of-experts (MoE) encoders could enable sustainable expansion to more modes, thereby increasing the model's capabilities. 
    \item \textbf{Application to Different Geoscience Tasks}: We have only tested our model on a single dataset, ERA5. To further assess the robustness of our approach, we could expand our solution to different geoscience datasets, incorporating more specific downstream tasks and yielding a more generalisable solution.
\end{enumerate}

\section{Data Availability Statement}
\vspace{3mm}

All code used to train and evaluate our models is available in a public repository: \url{https://github.com/nathanwbailey/SPARTA}. \\

All ERA5 data can be found from the official documentation: \url{https://weatherbench2.readthedocs.io/en/latest/data-guide.html}. Additionally, we include scripts in our repository to convert these datasets into PyTorch tensors for use in our code. 



\bibliographystyle{vancouver}
\footnotesize{
\bibliography{main.bib}}
\end{document}